\documentclass[letterpaper, 10pt]{article}
\usepackage{graphicx}
\usepackage{amsmath}
\usepackage{float}
\usepackage{amsfonts}
\usepackage[inner=0.75in, outer=0.75in, top=0.75in, bottom=1in, paperheight=11in, paperwidth=8.5in]{geometry}

\usepackage{epstopdf}
\usepackage{float}
\usepackage{amssymb}
\usepackage[utf8]{inputenc}
\usepackage[english]{babel}
\usepackage{amsthm}
\usepackage{color}
\usepackage{forest}

\usepackage{microtype}
\usepackage{url}
\usepackage{subfiles}
\usepackage{caption}
\usepackage{subcaption}
\usepackage{parskip}
\usepackage[section]{placeins}

\usepackage{tikz}
\usetikzlibrary{decorations.pathmorphing}
\usetikzlibrary{decorations.markings}
\usetikzlibrary{arrows.meta,bending}
\usetikzlibrary{arrows.meta}
\usepackage{bm}

\usepackage{xcolor}
\usepackage{framed}

\colorlet{shadecolor}{blue!20}

\usepackage[obeyFinal,textwidth=0.55in]{todonotes}
\usepackage[hidelinks]{hyperref}
\usepackage[capitalize]{cleveref}

\theoremstyle{definition}

\begin{document}	

\title{\bf On an application of graph neural networks in population based SHM}	
\author{G.\ Tsialiamanis$^1$, C.\ Mylonas$^2$, E.\ Chatzi$^2$, D.J.\ Wagg$^1$, N.\ Dervilis$^1$ \& K.\ Worden$^1$\\
        $^1$Dynamics Research Group, Department of Mechanical Engineering, University of Sheffield \\
        Mappin Street, Sheffield S1 3JD, UK\\
        $^2$ETH Zurich, Institute of Structural Engineering Stefano-Franscini-Platz 5, 8093 Zurich}
	\date{}
    \maketitle
	\thispagestyle{empty}

\section*{Abstract}
Attempts have been made recently in the field of population-based structural health monitoring (PBSHM), to transfer knowledge between SHM models of different structures. The attempts have been focussed on homogeneous and heterogeneous populations. A more general approach to transferring knowledge between structures, is by considering all plausible structures as points on a multidimensional base manifold and building a fibre bundle. The idea is quite powerful, since, a mapping between points in the base manifold and their fibres, the potential states of any arbitrary structure, can be learnt. A smaller scale problem, but still useful, is that of learning a specific point of every fibre, i.e. that corresponding to the undamaged state of structures  within a population. Under the framework of PBSHM, a data-driven approach to the aforementioned problem is developed. Structures are converted into graphs and inference is attempted within a population, using a graph neural network (GNN) algorithm. The algorithm solves a major problem existing in such applications. Structures comprise different sizes and are defined as abstract objects, thus attempting to perform inference within a heterogeneous population is not trivial. The proposed approach is tested in a simulated population of trusses. The goal of the application is to predict the first natural frequency of trusses of different sizes, across different environmental temperatures and having different bar member types. After training the GNN using part of the total population, it was tested on trusses that were not included in the training dataset. Results show that the accuracy of the regression is satisfactory even in structures with higher number of nodes and members than those used to train it.

\textbf{Key words: Structural health monitoring (SHM), machine learning, graph neural networks, fibre bundles, population-based structural health monitoring (PBSHM).}

\section{Introduction}
\label{sec:intro}
\textit{Structural health monitoring} (SHM) has developed a lot in recent years and is widely used with a view to maintaining structural health and to minimizing operation and maintenance costs. Many paths have been followed in order to perform SHM and they can be separated into two major categories, the \textit{physics-based} methods and the \textit{data-driven} methods. Physics-based methods for SHM - often used in \textit{condition monitoring} \cite{randall2011vibration} - are based on understanding the physics of the problem and defining quantities, whose monitoring can indicate existence of damage. Since one fully understands some structure's functionality, such methods use data but are mostly physics-informed and tend to be robust and universally applicable in specific types of structures (e.g. in rotating machinery). On the other hand, data-driven methods are based mainly on data acquired from structures \cite{farrar2012structural}. Following the latter scheme, classification and regression procedures can be performed, regarding the type of damage that might be present in a structure, the damage extent and even calculating the remaining useful life of a structure.

Data-driven methods can solve problems, even if one has no knowledge about the underlying physics of the problem. Under such a framework, \textit{machine learning} (ML), whose methods prove very powerful in learning from data, is employed. Different problems of SHM can be addressed using machine learning. Identification/localisation of damage type may be performed by training a classification algorithm on data acquired from some structure under different types of damage \cite{manson2003experimental}. Regression models can be used in order to predict the behaviour of an undamaged structure and potential deviations from the predicted behaviour may be a sign of damage \cite{fugate2001vibration}. Sequential models can be utilised to predict the evolution of damage on a structure. In any case, exact knowledge of the underlying physics is not a defining requirement. The models perform simply by learning relationships between input and output data in a black-box modelling scheme. Nevertheless, an essential requirement in order to apply any of these methods is the existence of data and the selection of the right features of the data.

Lack of sufficient (or labelled) data is probably the major drawback of most data-driven methods. Another major problem might be the definition of damage sensitive features, but most commonly this is solved using physical intuition. Regardless, if data are not available, none of these methods can be applied. Even in the trivial case of \textit{outlier detection}, data should be acquired from a structure during its operation in an undamaged state \cite{WORDEN2000}. The situation is less favourable for such methods in cases which require data from different states of a structure and under different types of damage.

To cope with situations of lack of data, \textit{transfer learning} has been deployed \cite{gardner2020application}. Using transfer learning one seeks to exploit models that exist for a specific task and domain and transfer knowledge, that they might have incorporated, in order to enhance the performance of a second model referring to a different task or domain. This strategy is usually followed when one has not sufficient data to train some machine learning model on a task. Therefore, knowledge that a model might have acquired through training on a different task is sought to be extracted and exploited. It is a straightforward approach in the discipline of \textit{computer vision} \cite{GabrielPuiCheongFung2006,AlMubaid2006}, to use \textit{neural networks}, where pre-trained filters from convolutional neural networks are used as feature extraction modules for a different image dataset and a different task.

A wide range of attempts to transfer knowledge between structures under an SHM framework have been performed according to the discipline of \textit{population-based structural health monitoring} (PBSHM). According to the newly-introduced discipline, inspired by the way medicine is applied on humans, structure populations are characterised either as \textit{homogeneous} or \textit{heterogeneous}. The first category refers to populations of structures with little differences between the members and the second to populations with individuals with significant differences. Attempts have been made in both cases in order to exploit existing knowledge about individuals of a population and transfer it to ones where sufficient data might not exist \cite{bull2020foundations, gosliga2020foundations, gardner2020foundations}. Specifically in \cite{gosliga2020foundations}, a conversion of structures into graphs is proposed, which will be also discussed and exploited herein. 

In the current work, a scheme of performing knowledge transfer slightly different to the ones presented before is proposed. The proposed scheme maps structures onto a base manifold and considers their normal condition data as fibres ``attached'' to each structure/point of the base manifold. The combination of every structure and every ``attached'' manifold (fibre) forms a \textit{fibre bundle}. The objective of transferring knowledge from one structure to another is considered equivalent to navigating through the structure manifold and inferring or predicting the fibre for points/structures, whose data are not available. In short, the transfer objective is reduced to interpolation between points of a base manifold and their respective fibre manifolds.

The layout of the paper is as follows. In Section \ref{sec:fibre_bundles}, a brief introduction to fibre bundles is provided; basic definitions are given and the framework is described according to the needs of PBSHM. Next, in Section \ref{sec:GNNs}, the machine learning algorithm - \textit{Graph Neural Networks} (GNNs) - that is used to perform the aforementioned interpolation between structures is introduced. In Section \ref{sec:experiments}, an application of the proposed procedure is presented in a simulated heterogeneous population of truss structures. The way the algorithm is applied is described, as well as the results of the application. Finally, in Section \ref{sec:conclusions}, a short discussion of the results is given.

\section{Fibre bundles}
\label{sec:fibre_bundles}

\subsection{Manifolds}
\label{sec:manifolds}
Manifolds are objects widely used both in physics and dynamics, as well as in machine learning. According to \cite{Schutz}, a manifold $M$ is a set of points for which, in an open neighbourhood around them, a $1-1$ continuous mapping exists onto $\mathbb{R}^{n}$ for some $n$. In the context of dynamics, manifolds can be used to explain phenomena like the oscillations of a pendulum. The configuration space of the pendulum is the circle $S^{1}$ and its phase space the cylinder $S^{1} \times \mathbb{R}$. In SHM and even more in data-driven SHM, manifolds are present in almost every application. The samples in datasets used in the discipline usually belong to manifolds and classification algorithms quite often attempt distinction between those, each one representing a different class. In cases of regression, a mapping between points on a manifold onto a second manifold is attempted.

\begin{figure}[htpb!]
    \centering
    \begin{tikzpicture}
        
        \draw[thick] (-2,0) -- (-2,2);
        
        \draw[thick] (-2, 2) -- (-2.2, 1.8);
        \draw[thick] (-2, 1.8) -- (-2.2, 1.6);
        \draw[thick] (-2, 1.6) -- (-2.2, 1.4);
        \draw[thick] (-2, 1.4) -- (-2.2, 1.2);
        \draw[thick] (-2, 1.2) -- (-2.2, 1.0);
        \draw[thick] (-2, 1.0) -- (-2.2, 0.8);
        \draw[thick] (-2, 0.8) -- (-2.2, 0.6);
        \draw[thick] (-2, 0.6) -- (-2.2, 0.4);
        \draw[thick] (-2, 0.4) -- (-2.2, 0.2);
        \draw[thick] (-2, 0.2) -- (-2.2, 0.0);
        
        \draw[thick, decoration={aspect=0.65, segment length=3mm,
             amplitude=0.2cm, coil}, decorate] (-2,1) --(0,1);
        
        \draw[thick] (0,0) rectangle (2,2) node[pos=.5] {$m_1, x_1, v_1$};
        
        \draw[thick, decoration={aspect=0.65, segment length=3mm,
             amplitude=0.2cm, coil}, decorate] (2,1) --(4,1);
             
        \draw[thick] (4,0) rectangle (6,2) node[pos=.5] {$m_2, x_2, v_2$};
        
        \draw[thick, decoration={aspect=0.65, segment length=3mm,
             amplitude=0.2cm, coil}, decorate] (6,1) --(8,1);
        
        \draw[thick] (8,0) rectangle (10,2) node[pos=.5] {$m_3, x_3, v_3$};
        
        \draw[->,thick] (1, 2) -- (1, 3) -- (2, 3);
        
        \node[] at (1.5, 3.5) {$F$};
        
        \node[] at (-1, 1.5) {$k_1$};
        \node[] at (3, 1.5) {$k_2$};
        \node[] at (7, 1.5) {$k_3$};
        
        \end{tikzpicture} 
    \caption{Three-degree-of-freedom mass spring system.}
    \label{fig:mass_spring}
\end{figure}
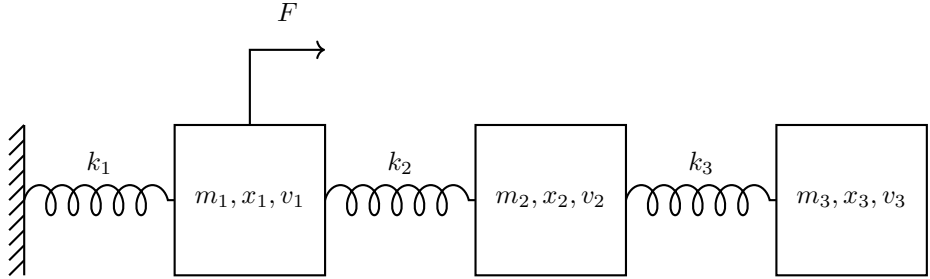

\begin{figure}[H]
    \centering
    \includegraphics[scale=0.35]{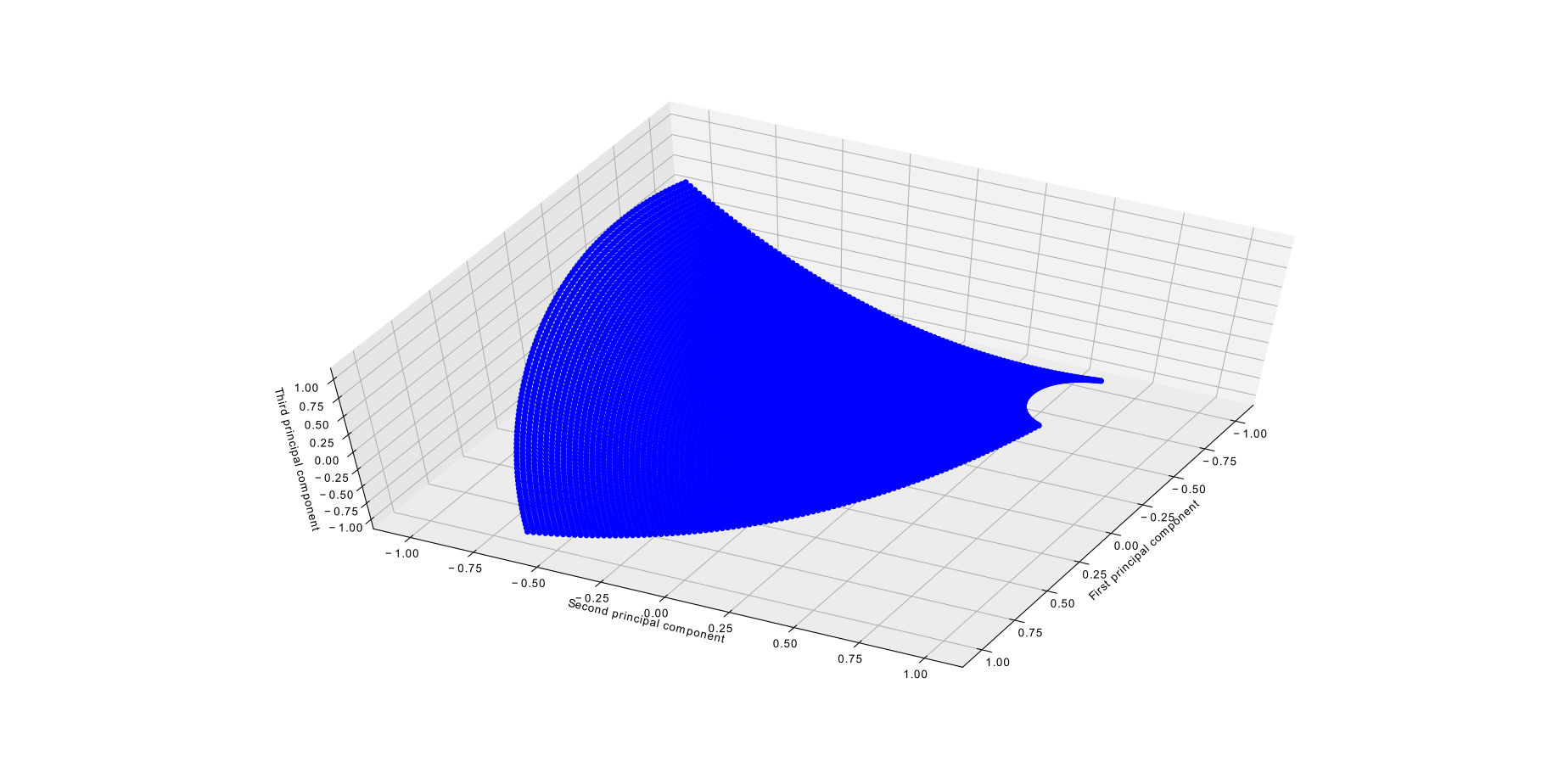}
    \caption{Manifold of points corresponding to the FRF of the lumped mass system under variations in the values of the stiffness of two springs.}
    \label{fig:PCA_dataset}
\end{figure}

In SHM, manifolds can explain the behaviour of a structure by collecting all points which represent its behaviour under different circumstances. For example, for a linear three-degree-of-freedom lumped mass system like the one shown in Figure \ref{fig:mass_spring}, the \textit{frequency response function} (FRF) of the first degree of freedom, under some random excitation $F$, is a single point in a multidimensional feature space (given that the FRF is comprised of discrete spectral lines). The dimension of the feature space is equal to the number of spectral lines of the FRF. If variations in stiffness are induced for springs $\#1$ and $\#2$ and again the same FRF is calculated, the FRF will correspond to a different point in the feature space. Collecting points for a set of variations of the two springs, a manifold is generated. This manifold could e.g. have the form of the manifold shown in Figure \ref{fig:PCA_dataset}. A \textit{principal component analysis} \cite{wold1987principal} (PCA) was performed in order to plot the multidimensional manifold here. If it is assumed that the specific structure operates under these variations of the stiffnesses, then this manifold characterises it and describes different states in which the structure might be observed. 

\subsection{Fiber bundles}
\label{sec:sub_fiber_bundles}

A fibre bundle is constructed by considering at first a \textit{base manifold} $M$. Subsequently, for every point in $M$, another manifold is defined. This latter manifold is called a \textit{fibre} ($F$). By collecting every fibre $F$ for every point $x \in M$, a total space $E$ is created. Of course, every fibre must be glued on the base manifold in order for movement in the base manifold to induce movement in the total space. Furthermore, a projection $\pi:E \to M$ is defined, projecting every point from the total space to the base manifold. This theory means that the inverse of $\pi$ must be multi-valued and it is a map from the base manifold to the fibre corresponding to each point, $\pi^{-1}(x) = F_{x} \in E$. A first requirement for the definition of the fibre bundle is that all fibres are \textit{homeomorphic}.

\begin{figure}[htbp!]
    \centering
     \begin{tikzpicture}
     \definecolor{blue1}{RGB}{93, 143, 218}
     \definecolor{teal}{RGB}{100, 225, 225}
     \draw[line width=0.2mm, blue1] (0.0, 0.0) to[out=30, in=150] (3.0, 0.5) to[out=330, in=210] (6.0, 1.0);
     \draw[line width=0.2mm, blue1] (6.0, 1.0) to[] (7.0, 3.0);
     \draw[line width=0.2mm, blue1] (1.0, 2.0) to[out=30, in=150] (3.0, 2.5) to[out=330, in=210] (7.0, 3.0);
     \draw[line width=0.2mm, blue1] (1.0, 2.0) to[] (0.0, 0.0);
     
     
     \draw[line width=0.2mm, blue1] (0.0, 4.0) to[] (0.0, 7.0);
     \draw[line width=0.2mm, blue1] (6.0, 4.5) to[] (6.0, 7.5);
     
     \draw[line width=0.2mm, blue1] (0.0, 7.0) to[] (1.0, 8.5);
     \draw[line width=0.2mm, blue1] (6.0, 7.5) to[] (7.0, 9.0);
     
     \draw[line width=0.2mm, blue1] (1.0, 8.5) to[out=60, in=150] (3.0, 8.75) to[out=330, in=240] (5.0, 8.75) to[out=60, in=130] (7.0, 9.0);
     
     \draw[line width=0.2mm, blue1] (6.0, 4.5) to[] (7.0, 6.0);
     \draw[line width=0.2mm, blue1] (7.0, 6.0) to[] (7.0, 9.0);
     
     \node[] (M) at (7.0, 2.0) {$M$};
     
     \node[] (E) at (7.6, 7.0) {$E$};
     
     \node[circle,color=blue1, fill=blue1, inner sep=0pt,minimum size=3pt] (fb1) at (2.0, 4.75) {};
     \node[circle,color=blue1, fill=blue1, inner sep=0pt,minimum size=3pt] (fb2) at (2.0, 7.75) {};
     \draw[line width=0.5mm, blue1] (fb1) to[] (fb2);
     \node[] (F) at (2.2, 5.75) {$F$};

     \node[circle,color=black, fill=blue1, inner sep=0pt,minimum size=3pt] (x) at (2.0, 1.0) {};
     \draw[-{>[scale=2.5, length=2, width=3]}, line width=0.4mm, color=teal] (x) to (fb1);
     
     \node[] (x_) at (2.3, 1.0) {$x$};
     \node[] (phi) at (2.5, 2.25) {$\pi^{-1}$};
     
     \draw[line width=0.2mm, blue1] (0.0, 4.0) to[out=60, in=150] (2.0, 4.25) to[out=330, in=240] (4.0, 4.25) to[out=60, in=130] (6.0, 4.5);
     \draw[line width=0.2mm, blue1] (0.0, 7.0) to[out=60, in=150] (2.0, 7.25) to[out=330, in=240] (4.0, 7.25) to[out=60, in=130] (6.0, 7.5);
     
     \end{tikzpicture}
    \caption{Schematic of basic objects in a fibre bundle.}
    \label{fig:l9_def_bundle}
\end{figure}
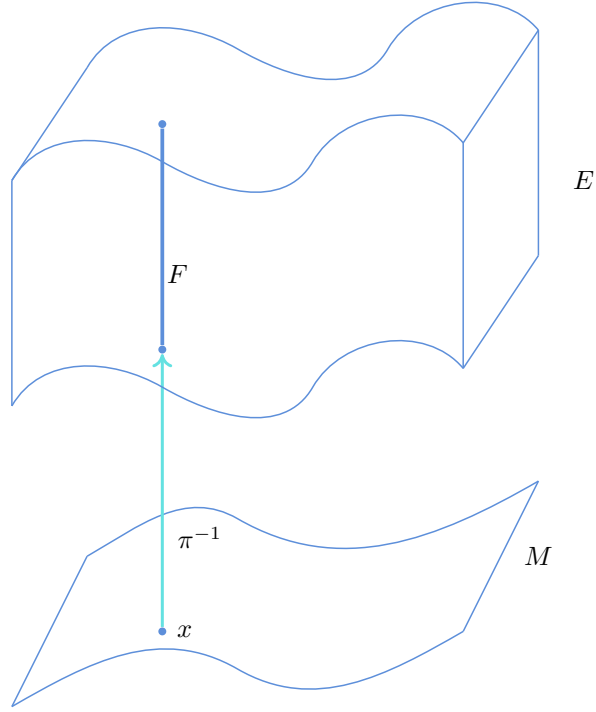

In total, the set of objects that comprise a fibre bundle are $\{M, E, \pi, F\}$ (example shown in Figure \ref{fig:l9_def_bundle}).  A trivial case would be for the total space to be simply the Cartesian product $M \times F$ but the fibre bundle concept is, and has to be, more flexible. It is allowed for $E$ to be only \textit{locally} homeomorphic to $M \times F$. Therefore, mappings from points in the manifold $M$ to their corresponding fibre $F_{i}$ can be defined locally in different domains $U_{i}$ of the manifold as $\phi_{i}: \pi^{-1}(U_{i}) \to U_{i} \times F$. This property is called \textit{local triviality} and it could prove very useful in using fibre bundles over a population of structures.

Another element of a fibre bundle that will be useful within the framework of PBSHM is that of a \textit{cross section} of the bundle. One might be interested in only one point out of every fibre instead of the whole manifolds. If every point of interest from every fibre is collected, they form a cross section of the fibre bundle. Such cross sections might have physical meaning, as it will be shown later, in the case of PBSHM since they may refer to characteristics of the structure during its undamaged state. The scheme is quite versatile and the features described by the fibres could refer to different temperatures or different levels of damage according to one's convenience.

\subsection{Fibre bundle over a population of structures}
\label{sec:FB_over_population}

In \cite{PBSHM6}, the way proposed in order to exploit the idea of a fibre bundle for PBSHM purposes, is first by defining a base manifold $M$ whose points correspond to different structures. Within the manifold, structures from the same population can be included, whether this is a homogeneous or a heterogeneous population. Consequently, for every structure a manifold shall be defined by collecting points, in some feature space, that represent their behaviour under every potential situation it might be observed into. Such an example was presented in the beginning of the current section, where the structure of Figure \ref{fig:mass_spring} was considered to be operating under variations of springs $\#1$ and $\#2$. By selecting as a feature of the structure, its FRF at the first degree of freedom and collecting points from every potential variation of the two springs' stiffness, the manifold of Figure \ref{fig:PCA_dataset} is formed. 

Similarly, for a population of structures, such manifolds may be collected for each individual within the population. Given that these manifolds correspond to undamaged states of each structure (this means that the variations are due to benign changing environmental conditions, nonlinearities, etc.), these manifolds are called the \textit{potential state manifolds} of the structures. If it is conveninient for some application, samples representing damage states may also be included in these manifolds. Having collected such manifolds, they can now be attached to each point of the base manifold, which was defined previously. Thus, a fibre bundle is defined having as a base manifold one whose points represent structures and each fibre defines their plausible feature characteristics.

Being optimistic, one can hope that these manifolds are glued together as they should be, and navigation in the manifold of structures will also mean gradual transformations of the corresponding fibres. However, there are cases that these manifolds might not even be homeomorphic. Still, the idea may be applied over populations of structures, rather than all potential structures in the world. It is expected that within a population, even a heterogeneous one, such homeomorphisms will be present and the desired knowledge transfer will be applicable. Considering the local triviality property of fibre bundles, a population of structures could belong to a specific domain $U_{i}$ (or a few neighbouring ones) of the whole fibre bundle of structures, making the application of the idea applicable within $U_{i}$. 

Rather than trying to estimate the whole manifold of potential states, one might intend approximating only a part of it, or even only a cross section. The most straightforward element of the whole bundle that one might be interested in is the \textit{normal condition cross section}. This is, of course, the collection of points that correspond to undamaged states of the structures. In case of linear structures, this shall be a single point if \textit{confounding influences} are not present. The proposed algorithm in the current paper in estimating the normal condition cross section also takes into account confounding influences such as environmental conditions and more specifically, temperature variations here.

\section{Graph neural networks}
\label{sec:GNNs}

\subsection{Introduction}
\label{sec:GNN_intro}
Machine learning has recently been one of the most powerful ways of solving SHM problems and in particular, data-driven ones. Under the framework of PBSHM, it has been employed before and a major part of the exploitation is via methods of knowledge transfer. Under the proposed framework of defining a fibre bundle over a manifold of structures, machine learning can be used once again in order to apply the idea of using data from structures to approximate regular normal condition characteristics for further structures. A major obstacle in applying some conventional methods is that there is not a straightforward mapping of structures onto points of a manifold. In order to overcome such issues, the machine learning algorithm used herein is the \textit{graph neural network algorithm} (GNN) \cite{Battaglia}.

GNN was introduced as a way of inducing \textit{inductive biases} regarding the structure of data in a machine learning algorithm. Inductive biases are essentially knowledge that one introduces to a machine learning algorithm and forces learning in a way that matches (according to one's knowledge) the physics of the underlying problem. Regularisation used in neural network training \cite{mc2001improving} can be considered as an inductive bias; by its use, smaller neural network weight values are favoured, resulting in smoother decision boundaries or regression curves. Another widely-used inductive bias is that of using convolutional layers in neural networks \cite{lawrence1997face}. This type of algorithm is sensitive to locality of features in an input image and this bias has been induced by the user by using convolutional layers. 

A representative example of problems that appear when traditional ML schemes are followed is also presented in \cite{Battaglia}. This problem further offers motivation for the development of the GNN algorithm. The problem is that of training a regression algorithm in order to approximate the center of mass of $n$ masses. A traditional neural network approach would require definition of an input vector for the network which would include the mass and the coordinates of every point (e.g. $\{m_1, x_1, m_2, x_2..., m_n, x_n\}$). Afterwards, the model would be trained given some input values and the calculated output coordinates of the center of mass. The algorithm will most certainly be efficient but it would fail under specific circumstances. The first case it would fail would be under permutation of the order of the input mass and coordinate values. Given that each mass and coordinate may come from a specific distribution, these distributions define the domain of the input variables for the neural network. The network therefore is only interpolating within these regions of the feature space and would most probably fail if values outside them are fed as input. The second case that most classic ML models fail is when the dimension of the inputs varies. The model mentioned before is trained and can only operate for systems of masses with exactly $n$ masses. Having more or less masses makes the application of the algorithm almost impossible. 

To deal with issues like this, but also induce the structure of the data into the training procedure, GNNs were developed and used. The algorithm, instead of using vectors as inputs and outputs, uses graphs. Such graphs have attributed nodes, edges and global features. Both inputs and outputs are graphs and the target values of the approximation can refer to node, edge or global values approximation. Using the algorithm, as it was developed in \cite{Battaglia} and will be described in next sections, information is exchanged between nodes and edges of the networks and so the computations take into account the connectivity of elements within the graph. More examples of benefits of the algorithm in various disciplines such as social network simulation \cite{perozzi2014deepwalk, Kipf}, biology \cite{dobson2003distinguishing}, chemistry \cite{ralaivola2005graph}, medicine \cite{zitnik2018modeling}, engineering \cite{wang2018videos} etc., are available in the literature.

The algorithm, being applicable to datasets comprised of graphs of various sizes, both in node and in edge terms, is a perfect candidate to be employed in solving the problem described before. Using such a data-driven algorithm, populations of structures of different numbers of degrees of freedom can be taken into account. Moreover, using the approach presented in \cite{gosliga2020foundations}, and converting structures into graphs, the problem of mapping them onto a manifold is bypassed. Concurrently, by using a structure-sensitive algorithm, further understanding of the underlying physics of the problem is expected, as is potential extrapolation capabilities outside the training domain of the model.

\subsection{Graph neural network elements}
\label{sec:GNN_elems}

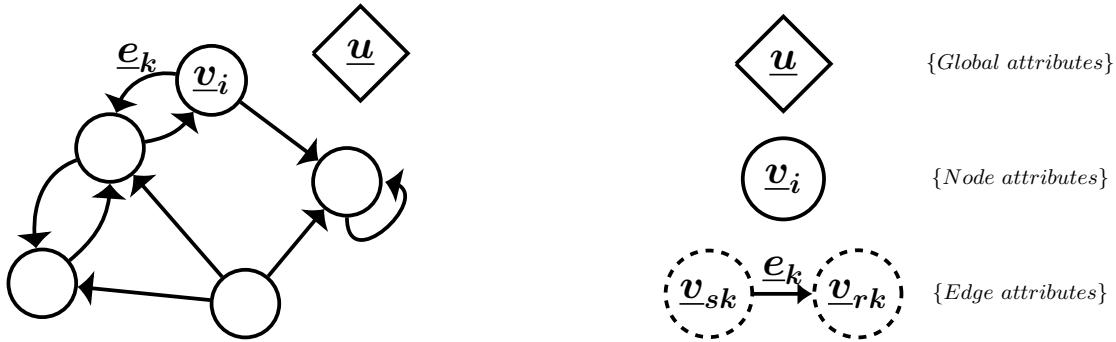
\begin{figure}[H]
    \centering
    \begin{subfigure}[b]{0.49\textwidth}
        \centering
        \begin{tikzpicture}[scale=0.9, every node/.style={scale=0.9}]
            \node[circle,draw, minimum size=1cm, line width=0.5mm] (A) at  (0, 0) {};
            \node[circle,draw, minimum size=1cm, line width=0.5mm] (B) at  (1, 2)  {};
            \draw[-{Latex[width=5mm, length=2mm]}, line width=0.5mm] (A) to[out=40, in=270] (B);
            \draw[-{Latex[width=5mm, length=2mm]}, line width=0.5mm] (B) to[out=200, in=100] (A);
            
            \node[circle, draw, minimum size=1cm, line width=0.5mm] (C) at  (3, -0.3) {};
            \node[circle, draw, minimum size=1cm, line width=0.5mm] (D) at  (2.5, 3) {\LARGE $\bm{{\underline{v}}_{i}}$};
            \node[circle, draw, minimum size=1cm, line width=0.5mm] (E) at  (4.5, 1.5) {};
            
            \draw[-{Latex[width=5mm, length=2mm]}, line width=0.5mm] (C) to (A);
            \draw[-{Latex[width=5mm, length=2mm]}, line width=0.5mm] (C) to (E);
            \draw[-{Latex[width=5mm, length=2mm]}, line width=0.5mm] (C) to (B);
            \draw[-{Latex[width=5mm, length=2mm]}, line width=0.5mm] (D) to (E);
            
            \draw[-{Latex[width=5mm, length=2mm]}, line width=0.5mm] (B) to[out=10, in=240] (D);
            \draw[-{Latex[width=5mm, length=2mm]}, line width=0.5mm] (D) to[out=170, in=80] node[above] {\LARGE $\bm{{\underline{e}}_{k}}$} (B);
            
            \draw[line width=0.5mm,-{Latex[width=5mm, length=2mm]},shorten >=1pt] (E) to [out=270,in=0,loop,looseness=3.4] (E);
            
            \draw[line width=0.5mm] (4, 3.4) -- (4.7, 2.7) -- (5.4, 3.4) -- (4.7, 4.1) -- (4, 3.4);
            \node[] at (4.7, 3.4) {\LARGE $\bm{{\underline{u}}}$};
        
        \end{tikzpicture}
    \end{subfigure}
    \begin{subfigure}[b]{0.49\textwidth}
        \centering
        \begin{tikzpicture}[scale=0.9, every node/.style={scale=0.9}]
            \draw[line width=0.5mm] (2, 4) -- (2.7, 3.3) -- (3.4, 4) -- (2.7, 4.7) -- (2, 4);
            \node[] at (2.7, 4) {\LARGE $\bm{{\underline{u}}}$};
            
            \node[circle, draw, minimum size=1.2cm, line width=0.5mm] (D) at  (2.7, 2.3) {\LARGE $\bm{{\underline{v}}_{i}}$};
            
            \node[circle, dashed, draw, minimum size=1cm, line width=0.5mm] (sender) at  (1.6, 0.6) {\LARGE $\bm{{\underline{v}}_{sk}}$};
            
            \node[circle, dashed, draw, minimum size=1cm, line width=0.5mm] (receiver) at  (3.8, 0.6) {\LARGE $\bm{{\underline{v}}_{rk}}$};
            
            \draw[-{Latex[width=3mm, length=2mm]}, line width=0.5mm] (sender) to node[above] {\LARGE $\bm{{\underline{e}}_{k}}$} (receiver);
            
            \node[] at (6.2, 4) {\small $\{Global\ attributes\}$};
            \node[] at (6.2, 2.3) {\small $\{Node\ attributes\}$};
            \node[] at (6.2, 0.6) {\small $\{Edge\ attributes\}$};
            
        \end{tikzpicture}
    \end{subfigure}
    \caption{General graph architecture.}
    \label{fig:general_graph}
\end{figure}

As already mentioned, GNNs use attributed graphs as inputs and outputs. Such a graph is shown in Figure \ref{fig:general_graph}. The elements of the graph are the edges, the nodes and the global attributes. Each element has an attribute vector assigned to it. Attribute vectors referring to same type of elements (i.e. nodes, edges or global) should have the same dimension. The edges are \textit{directed}, revealing the flow of information. Recurrent edges are allowed, making the algorithm even more versatile.

\begin{figure}[htbp!]
    \centering
    \begin{subfigure}[b]{0.80\textwidth}
        \centering
        \begin{tikzpicture}[scale=0.65, every node/.style={scale=0.65}]

            \draw[thick] (1, 1) circle (1.3cm) node {$Ground$};
            
            \draw[<->,thick] (2.3, 1) -- (3.3, 1);
            
            \draw[thick] (4.6, 1) circle (1.3cm) node {$[F_1, m_1, x_1, v_1]$};
            
            \draw[<->,thick] (5.9, 1) -- (6.9, 1);
            
            \draw[thick] (8.2, 1) circle (1.3cm) node {$[F_2, m_2, x_2, v_2]$};
            
            \draw[<->,thick] (9.5, 1) -- (10.5, 1);
            
            \draw[thick] (11.8, 1) circle (1.3cm) node {$[F_3, m_3, x_3, v_3]$};
            
            \node[] at (2.8, 1.5) {$[k_1]$};
            \node[] at (6.4, 1.5) {$[k_2]$};
            \node[] at (10.0, 1.5) {$[k_3]$};
            
            \draw[thick] (0, 4) -- (1.2, 2.8) -- (2.4, 4) -- (1.2, 5.2) -- (0, 4);
            
            \node[] at (1.2, 4) {$[T, K, U]$};
        
        \end{tikzpicture}
        \caption{Input mass-spring system graph.}
    \end{subfigure}
    \begin{subfigure}[b]{0.90\textwidth}
        \centering
        \begin{tikzpicture}[scale=0.65, every node/.style={scale=0.65}]

            \draw[thick] (1, 1) circle (1.3cm) node {$Ground$};
            
            \draw[<->,thick] (2.3, 1) -- (3.3, 1);
            
            \draw[thick] (4.6, 1) circle (1.3cm) node {$[x'_1, v'_1]$};
            
            \draw[<->,thick] (5.9, 1) -- (6.9, 1);
            
            \draw[thick] (8.2, 1) circle (1.3cm) node {$[x'_2, v'_2]$};
            
            \draw[<->,thick] (9.5, 1) -- (10.5, 1);
            
            \draw[thick] (11.8, 1) circle (1.3cm) node {$[x'_3, v'_3]$};
            
            \draw[thick] (0, 4) -- (1.2, 2.8) -- (2.4, 4) -- (1.2, 5.2) -- (0, 4);
        
        \end{tikzpicture}
        \caption{Output mass-spring system graph.}
    \end{subfigure}
    \caption{Graphs representing the input and the target output of the model, used to predict the displacements and velocities of the next time-step of the simulation.}
    \label{fig:graph_mass_spring_input_output}
\end{figure}
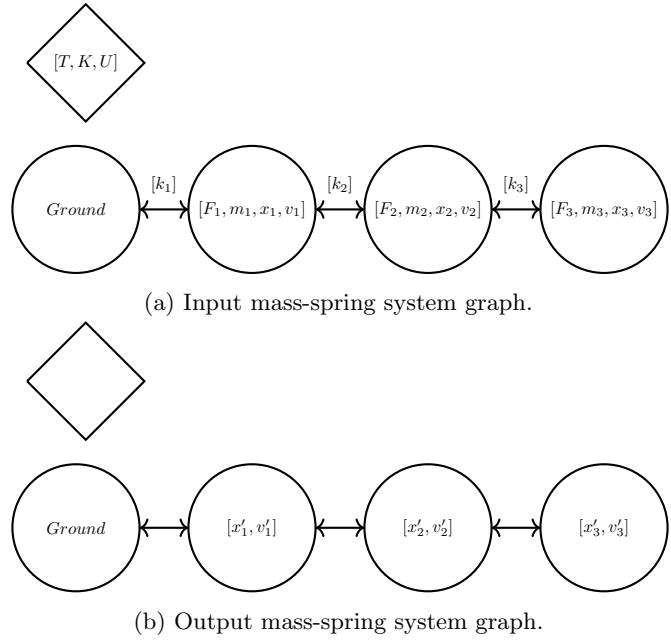

Both the input and the output of the algorithm are graphs, like the one shown above. If it is convenient for some application, elements of the graph may not have any attributes at all. Also, input and output attribute vectors do not need to be of the same dimension. An example about dynamics, could be the representation of a lumped mass system similar to the one shown in Figure \ref{fig:mass_spring}. The input graph could have as attributes to its nodes the mass, the displacement, the velocity and the force applied on each node, as edge features, the stiffness of each spring and as global attributes the temperature $T$ that the system is subjected into and the total kinetic $K$ and potential $U$ energy of the system. If one wants to predict the state (displacement and velocity) in the next time instant of the simulation using GNNs, the output graph would only have attributes in its nodes equal to these two quantities. The transformation of the system into input and output graphs for the purpose of applying the algorithm is shown in Figure \ref{fig:graph_mass_spring_input_output}.

\subsection{Graph neural network forward pass}
\label{sec:forward_pass}

Before defining the training procedure, it is worth dealing with the forward pass or the prediction procedure of the GNN algorithm. The procedure of predicting the output values for some given input graph is conducted in \textit{computational blocks}. Each computational block comprises three different steps, each one referring to updating of different types of attributes. The three steps are:

\begin{enumerate}
    \item the edge update,
    \item the node updates, and
    \item the global attribute updates.
\end{enumerate}

In each computational block, the number of which is pre-defined by the user, the updates are repeated in the order presented. The algorithm, being versatile, allows omitting any of the updates that one deems redundant. Such omissions could have underlying meaning and comprise another inductive bias introduced into the learning procedure.

\subsubsection{Edge update}
\label{sec:edge_update}
During the first step of the forward pass, the features of the edges are updated. The new attributes of some edge $e$ are updated into $\underline{e}'_{k}$ using the attributes of itself ($\underline{e}_{k}$), of the sender node ($\underline{v}_{sk}$), of the receiver node ($\underline{v}_{rk}$) as well as the global attributes of the graph ($\underline{u}$). The update equation is,

\begin{equation}
    \underline{e}'_{k} = \phi^{e}(\underline{e}_{k}, \underline{v}_{sk}, \underline{v}_{rk}, \underline{u})
    \label{eq:edge_update}
\end{equation}
where $\phi^{e}$ is a function learnt during training.

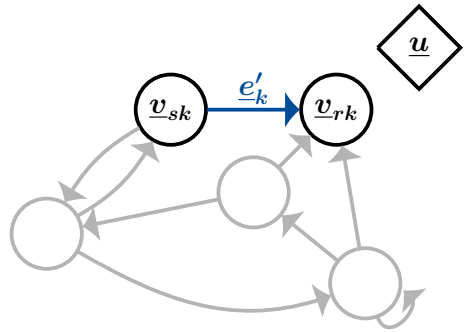
\begin{figure}[H]
    \centering
    \begin{tikzpicture}[scale=0.55, every node/.style={scale=0.55}]
        \definecolor{gray1}{RGB}{180, 180, 180}
        \definecolor{blue1}{RGB}{0, 76, 153}
        \node[circle,draw, minimum size=1.7cm, line width=0.5mm, gray1] (A) at  (0.5, 0) {};
        \node[circle,draw, minimum size=1.7cm, line width=0.5mm] (B) at  (3.5, 3) {\huge $\bm{{\underline{v}}_{sk}}$};
        \node[circle,draw, minimum size=1.7cm, line width=0.5mm] (C) at  (7.5, 3) {\huge $\bm{{\underline{v}}_{rk}}$};
        \node[circle,draw, minimum size=1.7cm, line width=0.5mm, gray1] (D) at  (5.5, 1.0) {};
        \node[circle,draw, minimum size=1.7cm, line width=0.5mm, gray1] (E) at  (8.2, -1.2) {};
        
        \draw[-{Latex[width=5mm, length=2mm]}, line width=0.5mm, gray1] (A) to[out=-30, in=200] (E);
        \draw[-{Latex[width=5mm, length=2mm]}, line width=0.5mm, gray1] (A) to[out=30, in=240] (B);
        \draw[-{Latex[width=5mm, length=2mm]}, line width=0.5mm, gray1] (B) to[out=210, in=60] (A);
        \draw[-{Latex[width=5mm, length=2mm]}, line width=0.5mm, blue1] (B) to node[above] {\huge $\bm{{\underline{e}}'_{k}}$} (C);
        
        \draw[line width=0.5mm,-{Latex[width=5mm, length=2mm]},shorten >=1pt, gray1] (E) to [out=290,in=-10,loop,looseness=2.4] (E);
        \draw[-{Latex[width=5mm, length=2mm]}, line width=0.5mm, gray1] (D) to (A);
        \draw[-{Latex[width=5mm, length=2mm]}, line width=0.5mm, gray1] (E) to (C);
        \draw[-{Latex[width=5mm, length=2mm]}, line width=0.5mm, gray1] (E) to (D);
        \draw[-{Latex[width=5mm, length=2mm]}, line width=0.5mm, gray1] (D) to (C);
        
        \draw[line width=0.5mm] (8.5, 4.5) -- (9.5, 3.5) -- (10.5, 4.5) -- (9.5, 5.5) -- (8.5, 4.5);
        \node[] at (9.5, 4.5) {\huge $\bm{\underline{u}}$};
    
    \end{tikzpicture}
    \caption{Edge update step.}
    \label{fig:edge_update}
\end{figure}

The procedure is depicted in Figure \ref{fig:edge_update}. This operation reveals even further the nature of the inductive bias used in the algorithm. The updating is local and depends on the connectivity of the network and of the specific edge. The update uses values only from the previous state of the graph and so it is also permutation invariant of the order of the updates.

\subsubsection{Node update}
\label{sec:node_update}

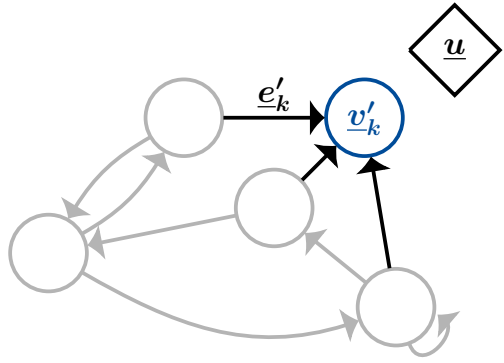
\begin{figure}[H]
    \centering
    \begin{tikzpicture}[scale=0.60, every node/.style={scale=0.60}]
        \definecolor{gray1}{RGB}{180, 180, 180}
        \definecolor{blue1}{RGB}{0, 76, 153}
        \node[circle,draw, minimum size=1.7cm, line width=0.5mm, gray1] (A) at  (0.5, 0) {};
        \node[circle,draw, minimum size=1.7cm, line width=0.5mm, gray1] (B) at  (3.5, 3) {};
        \node[circle,draw, minimum size=1.7cm, line width=0.5mm, blue1] (C) at  (7.5, 3) {\huge $\bm{\underline{v}'_{k}}$};
        \node[circle,draw, minimum size=1.7cm, line width=0.5mm, gray1] (D) at  (5.5, 1.0) {};
        \node[circle,draw, minimum size=1.7cm, line width=0.5mm, gray1] (E) at  (8.2, -1.2) {};
        
        \draw[-{Latex[width=5mm, length=2mm]}, line width=0.5mm, gray1] (A) to[out=-30, in=200] (E);
        \draw[-{Latex[width=5mm, length=2mm]}, line width=0.5mm, gray1] (A) to[out=30, in=240] (B);
        \draw[-{Latex[width=5mm, length=2mm]}, line width=0.5mm, gray1] (B) to[out=210, in=60] (A);
        \draw[-{Latex[width=5mm, length=2mm]}, line width=0.5mm] (B) to node[above] {\huge $\bm{{\underline{e}}'_{k}}$} (C);
        
        \draw[line width=0.5mm,-{Latex[width=5mm, length=2mm]},shorten >=1pt, gray1] (E) to [out=290,in=-10,loop,looseness=2.4] (E);
        \draw[-{Latex[width=5mm, length=2mm]}, line width=0.5mm, gray1] (D) to (A);
        \draw[-{Latex[width=5mm, length=2mm]}, line width=0.5mm] (E) to (C);
        \draw[-{Latex[width=5mm, length=2mm]}, line width=0.5mm, gray1] (E) to (D);
        \draw[-{Latex[width=5mm, length=2mm]}, line width=0.5mm] (D) to (C);
        
        \draw[line width=0.5mm] (8.5, 4.5) -- (9.5, 3.5) -- (10.5, 4.5) -- (9.5, 5.5) -- (8.5, 4.5);
        \node[] at (9.5, 4.5) {\huge $\bm{\underline{u}}$};
    
    \end{tikzpicture}
    \caption{Node update step.}
    \label{fig:node_update}
\end{figure}

The second step of the forward pass involves updating the features of the nodes. Having updated the edge features $E$, the new values $E'$ will be used in this step. The edges used are the ones which the current node is a receiver node. Due to the varying number of edges pointing at each node, an aggregative function $\rho^{e \to v}$ is used in order to unify the various inputs from the edges into one input vector. This function is usually a summation or an averaging function. Apart from the features of the edges that have the to-be-updated node as a receiver node, the features of the node itself are used and the global features of the graph. Similarly to before the update equation is,

\begin{equation}
    \underline{v}'_{k} = \phi^{v}(\rho^{e \to v}(E'), \underline{v}_k, \underline{u})
    \label{eq:node_update}
\end{equation}
where $\phi^{v}$ is also a function to be learnt. The procedure is schematically depicted in Figure \ref{fig:node_update}.

\subsubsection{Global update}
\label{sec:global_update}
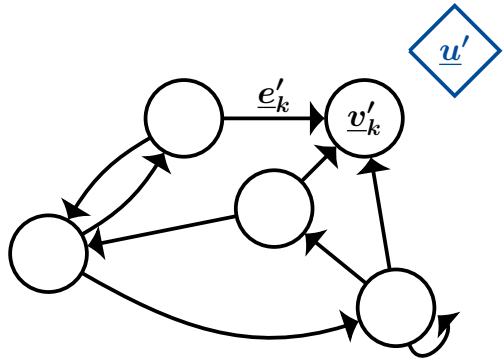
\begin{figure}[H]
    \centering
    \begin{tikzpicture}[scale=0.60, every node/.style={scale=0.60}]
        \definecolor{gray1}{RGB}{180, 180, 180}
        \definecolor{blue1}{RGB}{0, 76, 153}
        \node[circle,draw, minimum size=1.7cm, line width=0.5mm] (A) at  (0.5, 0) {};
        \node[circle,draw, minimum size=1.7cm, line width=0.5mm] (B) at  (3.5, 3) {};
        \node[circle,draw, minimum size=1.7cm, line width=0.5mm] (C) at  (7.5, 3) {\huge $\bm{\underline{v}'_{k}}$};
        \node[circle,draw, minimum size=1.7cm, line width=0.5mm] (D) at  (5.5, 1.0) {};
        \node[circle,draw, minimum size=1.7cm, line width=0.5mm] (E) at  (8.2, -1.2) {};
        
        \draw[-{Latex[width=5mm, length=2mm]}, line width=0.5mm] (A) to[out=-30, in=200] (E);
        \draw[-{Latex[width=5mm, length=2mm]}, line width=0.5mm] (A) to[out=30, in=240] (B);
        \draw[-{Latex[width=5mm, length=2mm]}, line width=0.5mm] (B) to[out=210, in=60] (A);
        \draw[-{Latex[width=5mm, length=2mm]}, line width=0.5mm] (B) to node[above] {\huge $\bm{{\underline{e}}'_{k}}$} (C);
        
        \draw[line width=0.5mm,-{Latex[width=5mm, length=2mm]},shorten >=1pt] (E) to [out=290,in=-10,loop,looseness=2.4] (E);
        \draw[-{Latex[width=5mm, length=2mm]}, line width=0.5mm] (D) to (A);
        \draw[-{Latex[width=5mm, length=2mm]}, line width=0.5mm] (E) to (C);
        \draw[-{Latex[width=5mm, length=2mm]}, line width=0.5mm] (E) to (D);
        \draw[-{Latex[width=5mm, length=2mm]}, line width=0.5mm] (D) to (C);
        
        \draw[line width=0.5mm, blue1] (8.5, 4.5) -- (9.5, 3.5) -- (10.5, 4.5) -- (9.5, 5.5) -- (8.5, 4.5);
        \node[blue1] at (9.5, 4.5) {\huge $\bm{\underline{u}'}$};
    
    \end{tikzpicture}
    \caption{Global update step.}
    \label{fig:global_update}
\end{figure}

The final updating step is shown in Figure \ref{fig:global_update}. As in the node update, a varying number of edges and nodes has to be taken into account, so similarly to the edge update, aggregative functions $\rho^{e \to u}$ and $\rho^{v \to u}$ are defined to deal with this issue. In this step, both the updated node ($V'$) and edge feature $E'$ vectors are used. The update equation is,

\begin{equation}
    \underline{u}' = \phi^{u}(\rho^{e \to u}(E'), \rho^{v \to u}(V'), \underline{u})
    \label{eq:global_update}
\end{equation}
where $\phi^{u}$ is a third function to be learnt.

\newpage

\subsection{Graph neural network training}
\label{sec:training}

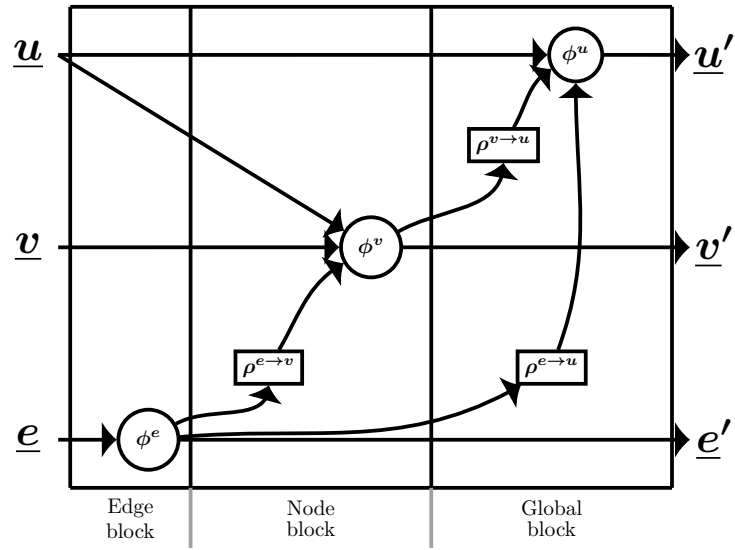
\begin{figure}[H]
    \centering
    \begin{tikzpicture}[scale=0.8, every node/.style={scale=0.8}]
        \definecolor{gray2}{RGB}{160, 160, 160}
        \draw[line width=0.5mm] (0, 0) to (0, 8);
        \draw[line width=0.5mm] (0, 8) to (10, 8);
        \draw[line width=0.5mm] (10, 8) to (10, 0);
        \draw[line width=0.5mm] (10, 0) to (0, 0);
        
        \node[] at (-0.7, 7.2) {\huge $\bm{\underline{u}}$};
        \node[] at (-0.7, 4.0) {\huge $\bm{\underline{v}}$};
        \node[] at (-0.7, 0.8) {\huge $\bm{\underline{e}}$};
        
        \node[] at (10.7, 7.2) {\huge $\bm{\underline{u}'}$};
        \node[] at (10.7, 4.0) {\huge $\bm{\underline{v}'}$};
        \node[] at (10.7, 0.8) {\huge $\bm{\underline{e}'}$};
        
        \node[circle,draw, minimum size=0.6cm, line width=0.5mm] (A) at  (8.4, 7.2) {$\bm{\phi^u}$};
        
        \draw[-{Latex[width=5mm, length=2mm]}, line width=0.5mm] (-0.2, 7.2) to (A);
        \draw[-{Latex[width=5mm, length=2mm]}, line width=0.5mm] (A) to (10.3, 7.2);
        
        \node[circle,draw, minimum size=1.0cm, line width=0.5mm] (B) at  (5.0, 4.0) {$\bm{\phi^v}$};
        \draw[-{Latex[width=5mm, length=2mm]}, line width=0.5mm] (-0.2, 7.2) to (B);
        \draw[-{Latex[width=5mm, length=2mm]}, line width=0.5mm] (-0.2, 4.0) to (B);
        \draw[-{Latex[width=5mm, length=2mm]}, line width=0.5mm] (B) to (10.3, 4.0);
        
        \node (rho3) at (7.2, 5.7) [draw,line width=0.5mm] {$\bm{\rho^{v \to u}}$};
        \draw[-{Latex[width=5mm, length=2mm]}, line width=0.5mm] (B) to[out=30, in=270] (rho3);
        \draw[-{Latex[width=5mm, length=2mm]}, line width=0.5mm] (rho3) to[out=60, in=210] (A);
        
        \node[circle,draw, minimum size=1.0cm, line width=0.5mm] (C) at  (1.3, 0.8) {$\bm{\phi^e}$};
        \draw[-{Latex[width=5mm, length=2mm]}, line width=0.5mm] (-0.2, 0.8) to (C);
        \draw[-{Latex[width=5mm, length=2mm]}, line width=0.5mm] (C) to (10.3, 0.8);
        
        \node (rho1) at (3.3, 2.0) [draw,line width=0.5mm] {$\bm{\rho^{e \to v}}$};
        \draw[-{Latex[width=5mm, length=2mm]}, line width=0.5mm] (C) to[out=30, in=270] (rho1);
        \draw[-{Latex[width=5mm, length=2mm]}, line width=0.5mm] (rho1) to[out=60, in=210] (B);
        
        \node (rho2) at (8.0, 2.0) [draw,line width=0.5mm] {$\bm{\rho^{e \to u}}$};
        \draw[-{Latex[width=5mm, length=2mm]}, line width=0.5mm] (C) to[out=5, in=210] (rho2);
        \draw[-{Latex[width=5mm, length=2mm]}, line width=0.5mm] (rho2) to[out=70, in=270] (A);
        
        \draw[line width=0.5mm] (2.0, 8.0) to (2.0, 0.0);
        \draw[line width=0.5mm, gray2] (2.0, 0.0) to (2.0, -1.0);
        \node[] at (1.0, -0.5) {\shortstack{Edge\\block}};
        
        \draw[line width=0.5mm] (6.0, 8.0) to (6.0, 0.0);
        \draw[line width=0.5mm, gray2] (6.0, 0.0) to (6.0, -1.0);
        \node[] at (4.0, -0.5) {\shortstack{Node\\block}};
        
        \node[] at (8.0, -0.5) {\shortstack{Global\\block}};
        
    \end{tikzpicture}
    \caption{Full computation block (motivated by \cite{Battaglia}).}
    \label{fig:full_computational_block}
\end{figure}

In practice, all $\phi$ functions mentioned are selected to be neural networks; doing so, the whole algorithm becomes trainable using back-propagation \cite{Bishop:1995:NNP:525960}. The full computational block can be seen in Figure \ref{fig:full_computational_block}. Every single one of the computations is differentiable and so back-propagation is applicable in order to tune the weights and biases of the neural network $\phi$ functions. 

\begin{figure}[H]
    \centering
    \begin{tikzpicture}[scale=0.8, every node/.style={scale=0.8}]
        \definecolor{gray2}{RGB}{160, 160, 160}
        \draw[line width=0.5mm] (0, 0) to (0, 8);
        \draw[line width=0.5mm] (0, 8) to (10, 8);
        \draw[line width=0.5mm] (10, 8) to (10, 0);
        \draw[line width=0.5mm] (10, 0) to (0, 0);
        
        \node[] at (-0.7, 7.2) {\huge $\bm{\underline{u}}$};
        \node[] at (-0.7, 4.0) {\huge $\bm{\underline{v}}$};
        \node[] at (-0.7, 0.8) {\huge $\bm{\underline{e}}$};
        
        \node[] at (10.7, 4.0) {\huge $\bm{\underline{v}'}$};

        \node[circle,draw, minimum size=1.0cm, line width=0.5mm] (B) at  (5.0, 4.0) {$\bm{\phi^v}$};
        \draw[-{Latex[width=5mm, length=2mm]}, line width=0.5mm] (-0.2, 7.2) to (B);
        \draw[-{Latex[width=5mm, length=2mm]}, line width=0.5mm] (-0.2, 4.0) to (B);
        \draw[-{Latex[width=5mm, length=2mm]}, line width=0.5mm] (B) to (10.3, 4.0);
        
        \node[circle,draw, minimum size=1.0cm, line width=0.5mm] (C) at  (1.3, 0.8) {$\bm{\phi^e}$};
        \draw[-{Latex[width=5mm, length=2mm]}, line width=0.5mm] (-0.2, 0.8) to (C);

        \node (rho1) at (3.3, 2.0) [draw,line width=0.5mm] {$\bm{\rho^{e \to v}}$};
        \draw[-{Latex[width=5mm, length=2mm]}, line width=0.5mm] (C) to[out=30, in=270] (rho1);
        \draw[-{Latex[width=5mm, length=2mm]}, line width=0.5mm] (rho1) to[out=60, in=210] (B);
        
        \draw[line width=0.5mm] (2.0, 8.0) to (2.0, 0.0);
        \draw[line width=0.5mm, gray2] (2.0, 0.0) to (2.0, -1.0);
        \node[] at (1.0, -0.5) {\shortstack{Edge\\block}};
        
        \draw[line width=0.5mm] (6.0, 8.0) to (6.0, 0.0);
        \draw[line width=0.5mm, gray2] (6.0, 0.0) to (6.0, -1.0);
        \node[] at (4.0, -0.5) {\shortstack{Node\\block}};
        
        \node[] at (8.0, -0.5) {\shortstack{Global\\block}};
        
    \end{tikzpicture}
    \caption{Reduced computation block.}
    \label{fig:reduced_comp_block}
\end{figure}
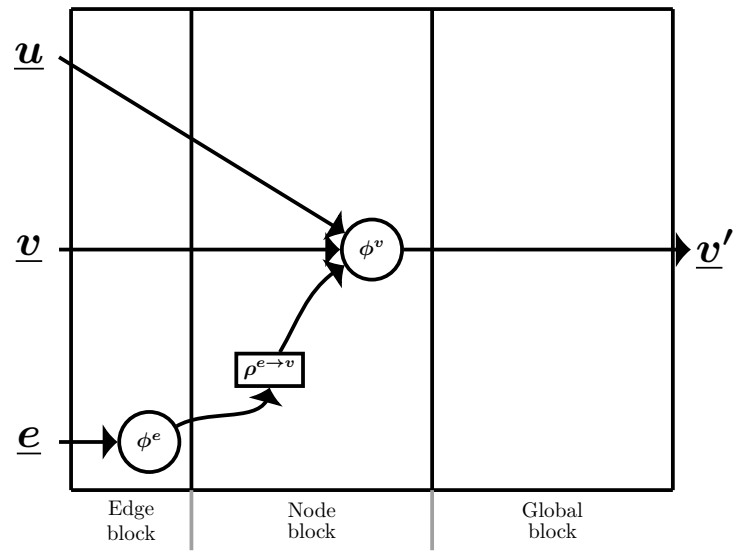

Even if several computational blocks are used until the final output graph is reached, the error of the output compared to the target values can be computed and back-propagated through gradient descent to the weights of the neural networks. The example presented in Figure \ref{fig:graph_mass_spring_input_output} offers a good chance to illustrate that the whole computational block is not always needed. In the specific example, the target outputs concern only node features. Therefore, the error will be calculated (most probably as a mean-square error) and back-propagated only through the node attributes ($\underline{\bm{v}}'$) of the output graph; in that case, the computational block would look like the one in Figure \ref{fig:reduced_comp_block}.

\section{Numerical experiments}
\label{sec:experiments}

In order to test the algorithm on a population of structures, a problem of defining the normal condition cross section within a population of trusses is considered. Trusses can be directly and unambiguously converted into \textit{attributed graphs}. Departing slightly from the original framework proposed in \cite{gosliga2020foundations}, according to which one would identify truss members (rods) as \textit{irreducible elements} (IE) and the joints as the connections, the nodes are considered as the IEs and the members as the connections. 

A planar truss model, like the one shown in Figure \ref{fig:planar_truss}, is naturally converted into an attributed graph. Nodes of the graph represent joints of the truss and their attributes are the $x$ and $y$ coordinates and another two variables with binary encoding, show whether the node is fixed in the $x$ or $y$ directions respectively. The physical attributes of the edges/rod members are the Young's modulus ($E$), the area of the member $A$ and its length $L$; in total, its stiffness is given by $K = \frac{EA}{L}$.

In need of a more practical and applicable way of encoding physical characteristics of the members, a \textit{categorical encoding} is proposed. A binary encoding of the characteristics of members proves more efficient in applications on existing structures. One might know that, within a population, some truss members are created using the same material but the exact properties under operational conditions might not be known. Therefore, by encoding each member according to its material and without knowing its exact properties, a more globally applicable algorithm is defined. Considering the truss from Figure \ref{fig:planar_truss}, one might assume that members on the top of the structure belong to a first class of members, members of the bottom to a second one and the members in the middle to a third. Following the encoding scheme described, the truss will be transformed in the attributed graph shown in Figure \ref{fig:planar_truss_model}. A global feature affecting the behaviour of the truss might be the temperature to which the structure is subjected.

\begin{figure}[H]
    \centering
        \begin{tikzpicture}[scale=0.8, every node/.style={scale=0.8}]
            \node[circle,draw=black, fill=black] (a) at (0, 0) {};
            
            \node[circle,draw=black, fill=black] (b) at (3, 0) {};
            
            \node[circle,draw=black, fill=black] (c) at (6, 0) {};
            
            \node[circle,draw=black, fill=black] (d) at (9, 0) {};
            
            \node[circle,draw=black, fill=black] (e) at (12, 0) {};
            
            \node[circle,draw=black, fill=black] (f) at (3, 4) {};
            
            \node[circle,draw=black, fill=black] (g) at (6, 4) {};
            
            \node[circle,draw=black, fill=black] (h) at (9, 4) {};
            
            \draw[line width=0.6mm, black] (a) -- (b);
            \draw[line width=0.6mm, black] (b) -- (c);
            \draw[line width=0.6mm, black] (c) -- (d);
            \draw[line width=0.6mm, black] (d) -- (e);
            \draw[line width=0.6mm, black] (a) -- (f);
            \draw[line width=0.6mm, black] (b) -- (f);
            \draw[line width=0.6mm, black] (f) -- (c);
            \draw[line width=0.6mm, black] (c) -- (g);
            \draw[line width=0.6mm, black] (c) -- (h);
            \draw[line width=0.6mm, black] (e) -- (h);
            \draw[line width=0.6mm, black] (d) -- (h);
            \draw[line width=0.6mm, black] (f) -- (g);
            \draw[line width=0.6mm, black] (g) -- (h);
            
            \node (v1) at (-.5, -0.9) {};
            \node (v2) at (0.5, -0.9) {};
            \fill[fill=black] (a.center)--(v1.center)--(v2.center);
            \draw[line width=0.3mm, black] (-.5, -0.9) -- ++(0.1, -0.1);
            \draw[line width=0.3mm, black] (-.4, -0.9) -- ++(0.1, -0.1);
            \draw[line width=0.3mm, black] (-.3, -0.9) -- ++(0.1, -0.1);
            \draw[line width=0.3mm, black] (-.2, -0.9) -- ++(0.1, -0.1);
            \draw[line width=0.3mm, black] (-.1, -0.9) -- ++(0.1, -0.1);
            \draw[line width=0.3mm, black] (0.0, -0.9) -- ++(0.1, -0.1);
            \draw[line width=0.3mm, black] (0.1, -0.9) -- ++(0.1, -0.1);
            \draw[line width=0.3mm, black] (0.2, -0.9) -- ++(0.1, -0.1);
            \draw[line width=0.3mm, black] (0.3, -0.9) -- ++(0.1, -0.1);
            \draw[line width=0.3mm, black] (0.4, -0.9) -- ++(0.1, -0.1);
            \draw[line width=0.3mm, black] (0.5, -0.9) -- ++(0.1, -0.1);
            
            \node (v4) at (11.5, -0.9) {};
            \node (v5) at (12.5, -0.9) {};
            \fill[fill=black] (e.center)--(v4.center)--(v5.center);
            \draw[line width=0.3mm, black] (11.5, -0.9) -- ++(0.1, -0.1);
            \draw[line width=0.3mm, black] (11.6, -0.9) -- ++(0.1, -0.1);
            \draw[line width=0.3mm, black] (11.7, -0.9) -- ++(0.1, -0.1);
            \draw[line width=0.3mm, black] (11.8, -0.9) -- ++(0.1, -0.1);
            \draw[line width=0.3mm, black] (11.9, -0.9) -- ++(0.1, -0.1);
            \draw[line width=0.3mm, black] (12.0, -0.9) -- ++(0.1, -0.1);
            \draw[line width=0.3mm, black] (12.1, -0.9) -- ++(0.1, -0.1);
            \draw[line width=0.3mm, black] (12.2, -0.9) -- ++(0.1, -0.1);
            \draw[line width=0.3mm, black] (12.3, -0.9) -- ++(0.1, -0.1);
            \draw[line width=0.3mm, black] (12.4, -0.9) -- ++(0.1, -0.1);
            \draw[line width=0.3mm, black] (12.5, -0.9) -- ++(0.1, -0.1);
            
        \end{tikzpicture}
    \caption{Simple planar truss model.}
    \label{fig:planar_truss}
\end{figure}
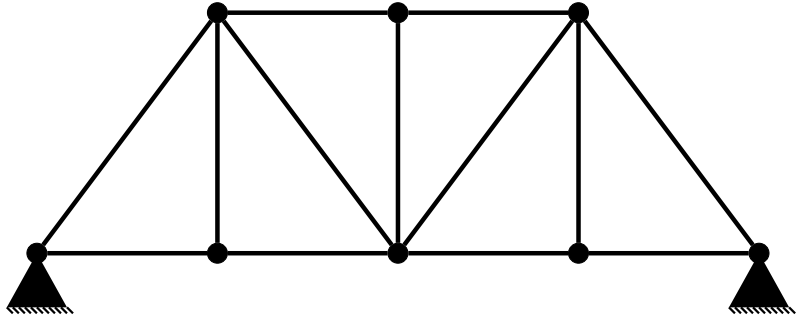

\begin{figure}[htbp!]
    \centering
    \begin{tikzpicture}[scale=0.8, every node/.style={scale=0.8}]

        \node[circle,draw=black, fill=white, label=below:\textcolor{blue}{$[x, y, 1, 1]$}] (a) at (0, 0) {};
        
        \node[circle,draw=black, fill=white, label=below:\textcolor{blue}{$[x, y, 0, 0]$}] (b) at (3, 0) {};
        
        \node[circle,draw=black, fill=white, label=below:\textcolor{blue}{$[x, y, 0, 0]$}] (c) at (6, 0) {};
        
        \node[circle,draw=black, fill=white, label=below:\textcolor{blue}{$[x, y, 0, 0]$}] (d) at (9, 0) {};
        
        \node[circle,draw=black, fill=white, label=below:\textcolor{blue}{$[x, y, 1, 1]$}] (e) at (12, 0) {};
        
        \node[circle,draw=black, fill=white, label=above:\textcolor{blue}{$[x, y, 0, 0]$}] (f) at (3, 4) {};
        
        \node[circle,draw=black, fill=white, label=above:\textcolor{blue}{$[x, y, 0, 0]$}] (g) at (6, 4) {};
        
        \node[circle,draw=black, fill=white, label=above:\textcolor{blue}{$[x, y, 0, 0]$}] (h) at (9, 4) {};
        
        \draw[thick] (a) -- (b) node[midway, below, text=orange] {$[0, 0, 1]$};
        \draw[thick] (b) -- (c) node[midway, below, text=orange] {$[0, 0, 1]$};
        \draw[thick] (c) -- (d) node[midway, below, text=orange] {$[0, 0, 1]$};
        \draw[thick] (d) -- (e) node[midway, below, text=orange] {$[0, 0, 1]$};
        \draw[thick] (a) -- (f) node[midway, above, sloped, text=orange] {$[0, 1, 0]$};
        \draw[thick] (f) -- (g) node[midway, above, sloped, text=orange] {$[1, 0, 0]$};
        \draw[thick] (g) -- (h) node[midway, above, sloped, text=orange] {$[1, 0, 0]$};
        \draw[thick] (b) -- (f) node[midway, above, sloped, text=orange] {$[0, 1, 0]$};
        \draw[thick] (f) -- (c) node[midway, above, sloped, text=orange] {$[0, 1, 0]$};
        \draw[thick] (c) -- (g) node[midway, above, sloped, text=orange] {$[0, 1, 0]$};
        \draw[thick] (c) -- (h) node[midway, above, sloped, text=orange] {$[0, 1, 0]$};
        \draw[thick] (e) -- (h) node[midway, above, sloped, text=orange] {$[0, 1, 0]$};
        \draw[thick] (d) -- (h) node[midway, above, sloped, text=orange] {$[0, 1, 0]$};
        
        \node[text=black!60!green] at (1.5, 5) {$[30]$};
    
    \end{tikzpicture}
    \caption{Network representation of truss shown in Figure \ref{fig:planar_truss}}
    \label{fig:planar_truss_model}
\end{figure}
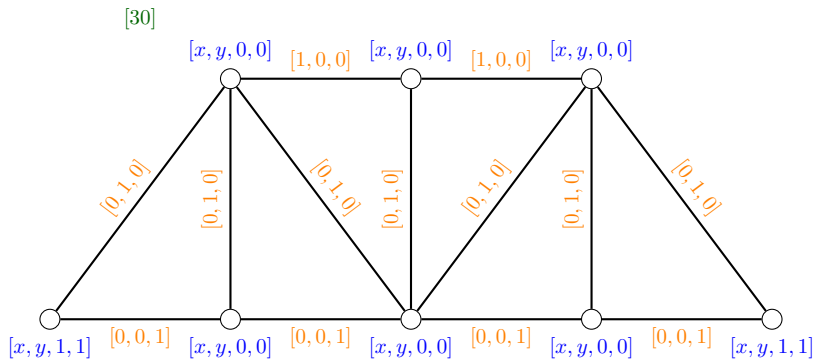

\begin{figure}[htbp!]
    \centering
    \includegraphics[scale=0.5]{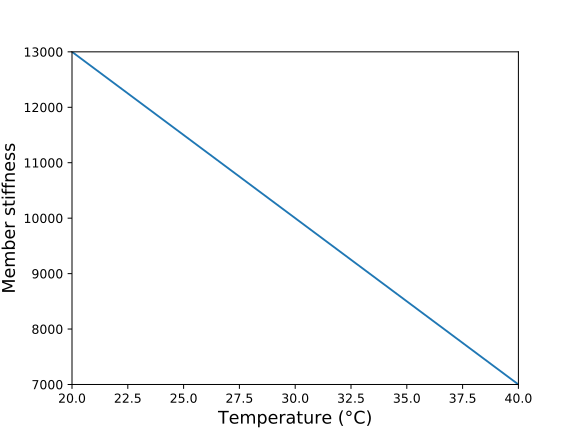}
    \caption{Linear relationship between temperature and EA of first type of members.}
    \label{fig:stiff_temp_1}
\end{figure}

\begin{figure}[htbp!]
    \centering
    \includegraphics[scale=0.5]{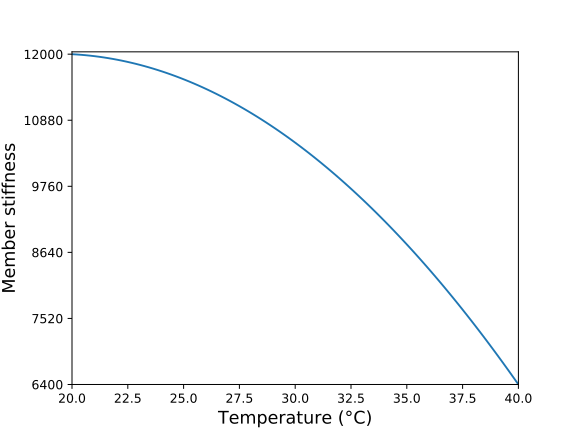}
    \caption{Nonlinear relationship between temperature and $EA$ of second type of members.}
    \label{fig:stiff_temp_2}
\end{figure}

Two case studies are performed and presented. The first one refers to a population of trusses assembled by members comprising the same Young's modulus $E$ and the same area $A$. The combined quantity $EA$ was equal to $10^{4}$ and the stiffness variation of the members was only due to their length $L$. In the second case study - the more complicated one here - two types of members and a temperature variation are considered. The first type of member has a linear relationship between its stiffness and temperature (shown in Figure \ref{fig:stiff_temp_1}), while the second has a nonlinear relationship between the two quantities (Figure \ref{fig:stiff_temp_2}).

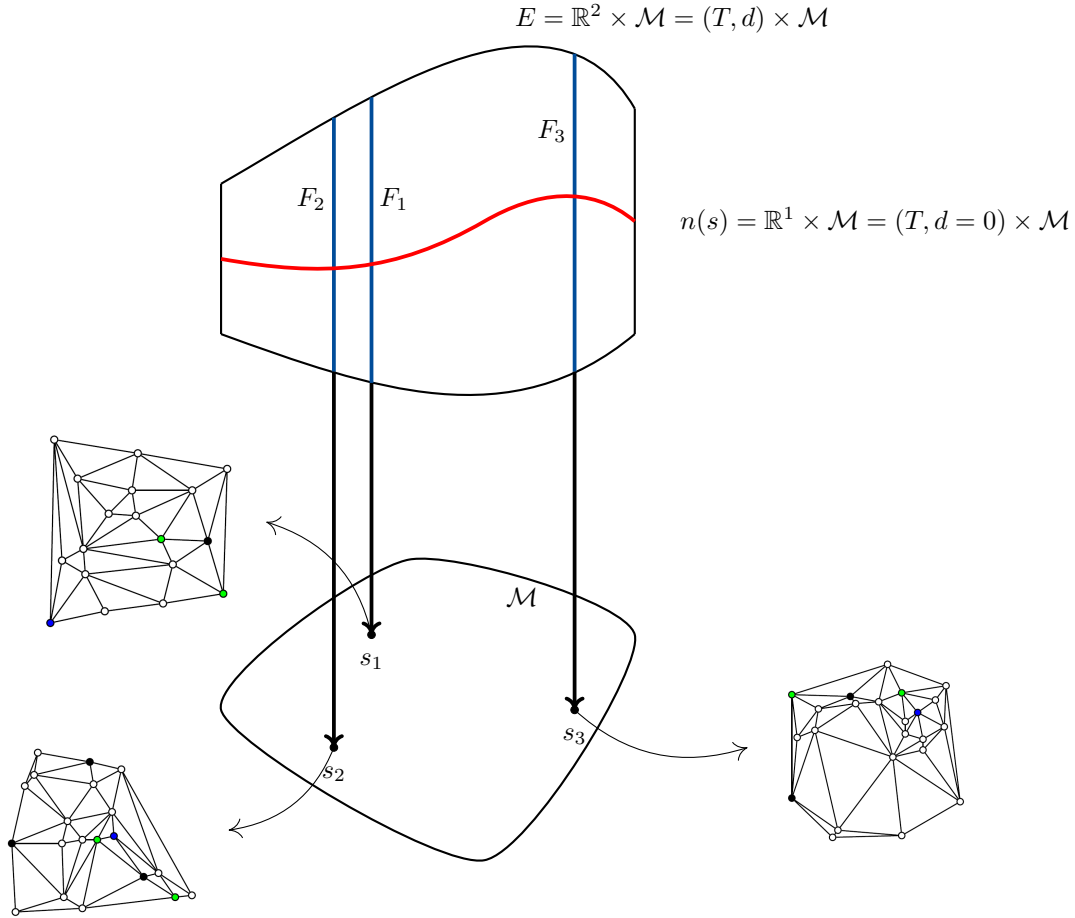
\begin{figure}[H]
    \centering
    \begin{tikzpicture}
        \draw[thick, smooth cycle, tension=0.4] plot coordinates{(2,2) (-0.5,0) (3,-2) (5,1)} node {};
        
        \definecolor{blue1}{RGB}{0, 76, 153}
        
        \path[-{>[scale=2.5, length=2, width=3]}] (4.2, 0) edge [bend right] node[left] {} (6.5, -.5);
        
        \path[-{>[scale=2.5, length=2, width=3]}] (1.0, -.5) edge [bend left] node[left] {} (-0.4, -1.6);
        
        \path[-{>[scale=2.5, length=2, width=3]}] (1.5, 1.0) edge [bend right] node[left] {} (0.1, 2.5);
        
        \draw[thick] (-0.5, 5.0) to[out=-20,in=-140] (5.0, 5.0);
        \draw[thick] (-0.5, 7.0) to[out=30,in=120] (5.0, 8.0);
        \draw[thick] (-0.5, 7.0) -- (-0.5, 5.0);
        \draw[thick] (5.0, 5.0) -- (5.0, 8.0);
        
        \draw[line width=0.5mm, blue1] (1.5, 8.15) to (1.5, 4.35);
        \draw[-{>[scale=2.5, length=2, width=3]}, line width=0.5mm] (1.5, 4.35) to (1.5, 1.0);
        \draw[black, fill=black] (1.5, 1.0) circle [radius=0.05cm];
        \node[label=below:{$s_{1}$}] at (1.5, 1.0) {};
        \node[] at (1.8, 6.8) {$F_1$};
        
        \draw[line width=0.5mm, blue1] (1.0, 7.88) to (1.0, 4.5);
        \draw[-{>[scale=2.5, length=2, width=3]}, line width=0.5mm] (1.0, 4.5) to (1.0, -.5);
        \draw[black, fill=black] (1.0, -.5) circle [radius=0.05cm] ;
        \node[label=below:{$s_{2}$}] at (1.0, -.5) {};
        \node[] at (0.7, 6.8) {$F_2$};
        
        \draw[line width=0.5mm, blue1] (4.2, 8.73) to (4.2, 4.5);
        \draw[-{>[scale=2.5, length=2, width=3]}, line width=0.5mm] (4.2, 4.5) to (4.2, 0);
        \draw[black, fill=black] (4.2, 0) circle [radius=0.05cm] ;
        \node[label=below:{$s_{3}$}] at (4.2, 0) {};
        \node[] at (3.9, 7.7) {$F_3$};
        
        \draw[line width=0.5mm, red] (-0.5, 6.0) to[out=-10,in=-150] (3.0, 6.5) to[out=30,in=140] (5.0, 6.5);
        
        \node[inner sep=0pt] (truss_1) at (-1.6, 2.4) {\includegraphics[width=.15\textwidth]{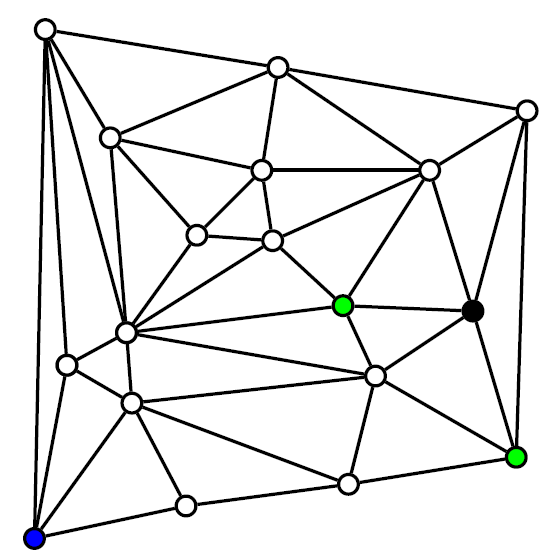}};
        
        \node[inner sep=0pt] (truss_1) at (-2.1, -1.6) {\includegraphics[width=.15\textwidth]{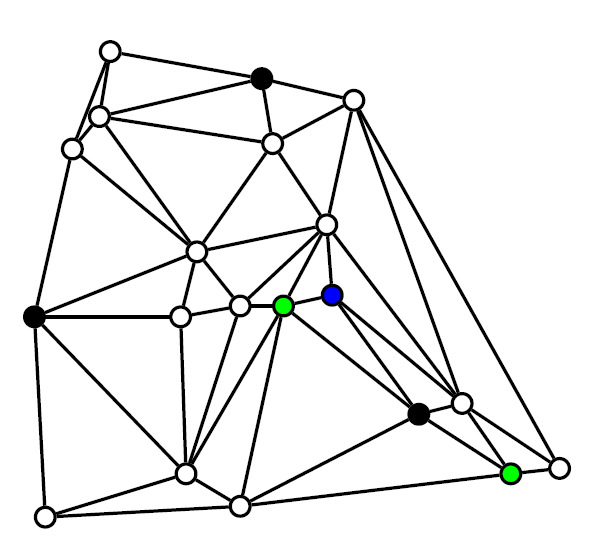}};
        
        \node[inner sep=0pt] (truss_1) at (8.2, -0.5) {\includegraphics[width=.15\textwidth]{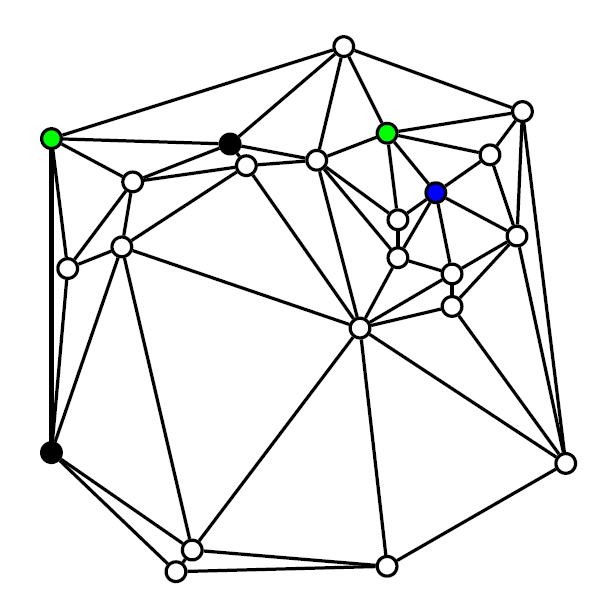}};
        
        \node[] at (3.5, 1.5) {$\mathcal{M}$};
        \node[] at (5.5, 9.2) {$E = \mathbb{R}^{2} \times \mathcal{M} = (T, d) \times \mathcal{M}$};
        
        \node[] at (8.2, 6.5) {$n(s) = \mathbb{R}^{1} \times \mathcal{M} = (T, d=0) \times \mathcal{M}$};
        
    \end{tikzpicture}
    \caption{Abstract scheme of the application. Nodes within $\mathcal{M}$ represent random trusses from within the population (three examples shown here: black nodes represent fully fixed nodes, green nodes are fixed in the $x$ direction, blue are fixed in the $y$ direction and in white are free (pinned) nodes). A fibre $F_i$ for each structure $s_i$ is schematically shown as a blue line. It is parametrised by all potential first natural frequencies of the structure for varying temperature T and damage coefficient $d$. The algorithm here has approximated the normal condition cross section $n(s)$ where damage $d$ is equal to $0$ and for any potential temperature $T$.} 
    \label{fig:schematic representation}

\end{figure}

In every case study, the population was created by randomly generating two-dimensional trusses with number of nodes in the interval $[10, 40]$. Delaunay triangulation \cite{Delaunay1934} was performed in order for the trusses to be realistic. Any temperature used was randomly sampled in the interval $[20, 40]$. The features of the nodes are according to Figure \ref{fig:planar_truss_model}, the coordinates $x$ and $y$ and two variables for binary encoding of the boundary conditions. The connections, representing the truss members had features equal to the length, as well as the $sine$ and $cosine$ of the angle of each member, since it was noted that this way training was facilitated. The feature whose approximation was sought was the first natural frequency of every truss. All case studies are therefore, an approximation of the normal condition cross-section of the plausible first natural frequencies of the simulated population. A schematic approach of the attempted application is shown in Figure \ref{fig:schematic representation}

\newpage

\subsection{Case study one}
\label{sec:case_study_1}

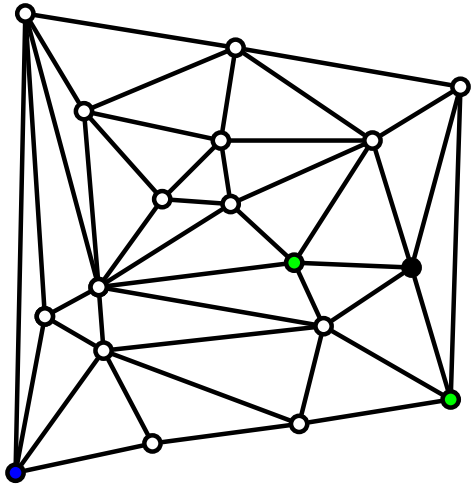
\begin{figure}[htbp!]
    \centering
    \begin{tikzpicture}[scale=0.65, every node/.style={scale=0.65}]
        \node[circle, line width=0.6mm, draw=black, fill=white] (0) at (9.6, 8.1) {};
        
        \node[circle, line width=0.6mm,draw=black, fill=green] (1) at (9.4, 1.7) {};
        
        \node[circle, line width=0.6mm,draw=black, fill=white] (2) at (0.7, 9.6) {};
        
        \node[circle, line width=0.6mm,draw=black, fill=white] (3) at (6.3, 1.2) {};
        
        \node[circle, line width=0.6mm,draw=black, fill=white] (4) at (2.3, 2.7) {};
        
        \node[circle, line width=0.6mm,draw=black, fill=white] (5) at (6.8, 3.2) {};
        
        \node[circle, line width=0.6mm,draw=black, fill=white] (6) at (3.5, 5.8) {};
        
        \node[circle, line width=0.6mm,draw=black, fill=white] (7) at (2.2, 4.0) {};
        
        \node[circle, line width=0.6mm,draw=black, fill=white] (8) at (2.2, 4.0) {};
        
        \node[circle, line width=0.6mm,draw=black, fill=green] (9) at (6.2, 4.5) {};
        
        \node[circle, line width=0.6mm,draw=black, fill=white] (10) at (4.7, 7.0) {};
        
        \node[circle, line width=0.6mm,draw=black, fill=black] (11) at (8.6, 4.4) {};
        
        \node[circle, line width=0.6mm,draw=black, fill=white] (12) at (3.3, 0.8) {};
        
        \node[circle, line width=0.6mm,draw=black, fill=white] (13) at (1.9, 7.6) {};
        
        \node[circle, line width=0.6mm,draw=black, fill=white] (14) at (4.9, 5.7) {};
        
        \node[circle, line width=0.6mm,draw=black, fill=white] (15) at (7.8, 7.0) {};
        
        \node[circle, line width=0.6mm,draw=black, fill=blue] (16) at (0.5, 0.2) {};
        
        \node[circle, line width=0.6mm,draw=black, fill=white] (17) at (1.1, 3.4) {};
        
        \node[circle, line width=0.6mm,draw=black, fill=white] (18) at (1.1, 3.4) {};
        
        \node[circle, line width=0.6mm,draw=black, fill=white] (19) at (5.0, 8.9) {};
        
        \draw[line width=0.6mm] (16) -- (12);
        \draw[line width=0.6mm] (16) -- (4);
        \draw[line width=0.6mm] (16) -- (18);
        \draw[line width=0.6mm] (16) -- (2);
        \draw[line width=0.6mm] (18) -- (4);
        \draw[line width=0.6mm] (18) -- (8);
        \draw[line width=0.6mm] (18) -- (2);
        \draw[line width=0.6mm] (4) -- (12);
        \draw[line width=0.6mm] (4) -- (3);
        \draw[line width=0.6mm] (12) -- (3);
        \draw[line width=0.6mm] (3) -- (1);
        \draw[line width=0.6mm] (3) -- (5);
        \draw[line width=0.6mm] (1) -- (5);
        \draw[line width=0.6mm] (4) -- (5);
        \draw[line width=0.6mm] (4) -- (8);
        \draw[line width=0.6mm] (8) -- (5);
        \draw[line width=0.6mm] (8) -- (13);
        \draw[line width=0.6mm] (8) -- (2);
        \draw[line width=0.6mm] (2) -- (13);
        \draw[line width=0.6mm] (2) -- (19);
        \draw[line width=0.6mm] (13) -- (19);
        \draw[line width=0.6mm] (13) -- (6);
        \draw[line width=0.6mm] (13) -- (10);
        \draw[line width=0.6mm] (10) -- (19);
        \draw[line width=0.6mm] (6) -- (10);
        \draw[line width=0.6mm] (8) -- (6);
        \draw[line width=0.6mm] (6) -- (14);
        \draw[line width=0.6mm] (14) -- (10);
        \draw[line width=0.6mm] (14) -- (8);
        \draw[line width=0.6mm] (9) -- (8);
        \draw[line width=0.6mm] (14) -- (9);
        \draw[line width=0.6mm] (9) -- (5);
        \draw[line width=0.6mm] (9) -- (11);
        \draw[line width=0.6mm] (5) -- (11);
        \draw[line width=0.6mm] (1) -- (11);
        \draw[line width=0.6mm] (14) -- (15);
        \draw[line width=0.6mm] (9) -- (15);
        \draw[line width=0.6mm] (15) -- (11);
        \draw[line width=0.6mm] (10) -- (15);
        \draw[line width=0.6mm] (19) -- (15);
        \draw[line width=0.6mm] (19) -- (0);
        \draw[line width=0.6mm] (15) -- (0);
        \draw[line width=0.6mm] (11) -- (0);
        \draw[line width=0.6mm] (1) -- (0);
    \end{tikzpicture}
    \caption{Random truss sampled from the dataset. Black nodes represent nodes fixed in both directions, green nodes are fixed only in the $x$ direction and blue ones are fixed only in $y$ direction.}
    \label{fig:random_truss}
\end{figure}

In the first case study, as in all of them, three datasets were generated for the purposes of cross-validation (training, validation and testing). Each one was comprised of $16000$ truss configurations, a sample of which is shown in Figure \ref{fig:random_truss}. After testing different numbers of computational blocks, it was established through cross-validation that three computational blocks were the most efficient.

Due to the high computational cost of the algorithm, a cross-validation was performed but not in an exhaustive sense. Different values for the hyperparameters of the models were tried in a range of their corresponding values. The whole procedure was closer to a trial and error one, having as model selection criterion the validation accuracy of the algorithm. An example of a hyperparameter whose exhaustive cross-validation procedure would result in hours of training even on a powerful desktop computer, is the architecture of the neural networks representing the $\phi$ functions. If for the set of the three neural networks and for every computational block, a large number of hidden layers number and hidden layer size were to be tested, the computational burden would probably be unbearable for a research group without appropriately powerfull computers. 

The results are evaluated according to a standard least-squares loss function. The error, in order to be properly evaluated, was a \textit{normalised mean-square error} (NMSE) defined by the equation,

\begin{equation}
    \centering
    \label{eq:NMSE}
    NMSE = \frac{100}{N\sigma_{f_1}^{2}}\sum_{i=1}^{N}(\hat{\omega}_{i} - \omega_{i})^{2}
\end{equation}
where $\omega_{i}$ is the target value of the first natural frequency and $\hat{\omega}_{i}$ is the estimated value from the algorithm. $N$ is the number of samples used to compute the NMSE and $\sigma_{\omega}^{2}$ is the variance of $\omega$ within the dataset. The NMSE is a useful metric, since values close to $100\%$ indicate that the model is predicting values close to the mean value of the target quantity, while smaller values indicate that the model has achieved a good fit on the data.

\subsubsection{Mean aggregative function}
\label{sec:mean_aggr_func}

Following the original approach to GNN and using as aggregative function $\rho$ a mean function, different numbers of computational blocks (CBs) were tried and as already mentioned, three CBs were the optimal. Regarding the architecture of each neural network used in the algorithm, a coarse search was performed using 10 random initialisation for every architecture tested. Interestingly, it was noted that the size of the neural networks of each CB should be larger than the previous CB in order to achieve acceptable accuracy. Therefore, the strategy followed was to gradually increase the sizes of the first CB neural networks and also impose proportional increase in the size of the later CBs.

The sizes of the networks in the first CB were tested using 20 to 600 units with an increment of about 20. Following the aforementioned scheme, the architectures which yield the best results are shown in Table \ref{Tab:GNN_arch_1}. The numbers represent in sequence the size of the neural network layers. For example $3, 64, 32$ represents a neural network with a three-node input layer, a $64$-node hidden layer and a $32$-node output layer.

\begin{table}[!htbp]
	\centering
	\begin{tabular}{ |p{3cm}||p{3cm}|p{3cm}|p{3cm}|}
		\hline
		 & First CB & Second CB & Third CB\\
		\hline
		Edge NN     & 3, 64, 32 & 132, 80, 50   & 290, 180, 80 \\
		\hline
		Vertices NN & 4, 72, 50  & 100, 100, 70  & 250, 300, 72 \\
		\hline
		Global NN   & -  & 120, 200, 100  & 252, 300, 72, 1 \\
		\hline
	\end{tabular}
	\caption{Sizes of neural networks used in the GNN model with mean aggregative function, first experiment.}
	\label{Tab:GNN_arch_1}
\end{table}

\begin{figure}[H]
    \centering
    \includegraphics[scale=0.7]{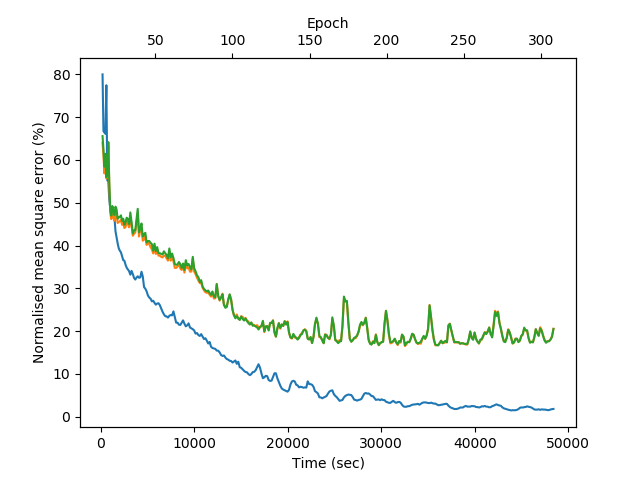}
    \caption{Training history of model using the summation aggregative function, training (blue), validation (orange) and testing (green) datasets, first experiment.}
    \label{fig:sum_aggr_training_history}
\end{figure}

The NMSE of the model was $1.55\%$ on the training data but it was substantially higher for the validation and testing sets, $17.16\%$ and $17.20\%$ respectively. The training history diagram is shown in Figure \ref{fig:sum_aggr_training_history}. The consistency of the error in the validation and testing datasets is encouraging the belief that the algorithm has incorporated a great part of the underlying physics of the problem. The difference in the error can be explained as the model specialising in performing on the training dataset. The total space of plausible trusses given the way they were assembled is really large and even such a big dataset may not be representative of the whole space. Even though, the performance on the validation and testing datasets is acceptable. 

\subsubsection{Augmented aggregative function}
\label{sec:augm_aggr_func}

Being concerned about the information that gets lost due to the use of a summation as an aggregative function, an alternative one is chosen in order to let more information flow through in these parts of the computations. As an alternative, a combination of an averaging and a variance computation is considered. An augmented output vector results from the computation. The augmented vector provides more information about the distribution of the quantities to be passed in the latter stages of the computation.

\begin{figure}[H]
    \centering
    \includegraphics[scale=0.7]{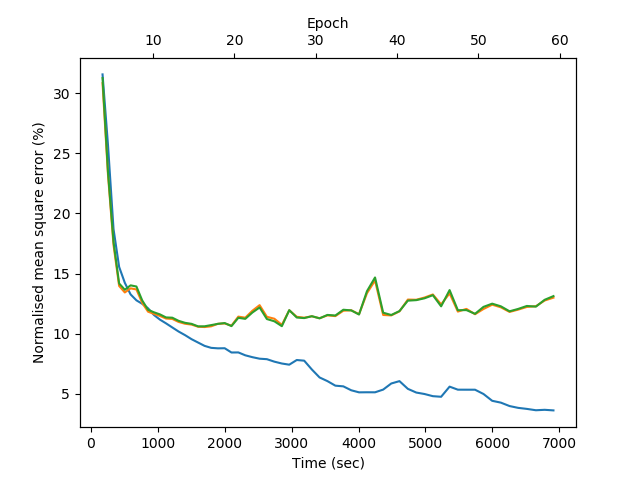}
    \caption{Training history of model using the augmented aggregative function, training (blue), validation (orange) and testing (green) datasets: first case study.}
    \label{fig:augm_aggr_training_history_1}
\end{figure}

The first case study was repeated and this time the neural network architectures are shown in Table \ref{Tab:GNN_arch_1_augmented}. It is worth noting that due to the augmentation of the aggregative functions, some input and hidden layers have significantly more units than in the previous case. The training history this time is shown in Figure \ref{fig:augm_aggr_training_history_1}. The NMSE is considerably lower for both the validation and training datasets ($10.54\%$ and $10.61\%$ respectively). Another advantage of using the augmented aggregative function was that the lowest error was achieved much faster. In the previous case it was achieved after 22000 seconds (6.1 hours) of training, while now it was found after 1500 seconds (25 minutes). As with the summation aggregation, the error on the training dataset approaches zero after hours of training. 

\begin{table}[!htbp]
	\centering
	\begin{tabular}{ |p{3cm}||p{3cm}|p{3cm}|p{3cm}|}
		\hline
		 & First CB & Second CB & Third CB\\
		\hline
		Edge NN     & 3, 64, 32 & 132, 80, 50   & 290, 180, 80 \\
		\hline
		Vertices NN & 4, 72, 50  & 150, 100, 70  & 330, 300, 72 \\
		\hline
		Global NN   & -  & 240, 200, 100  & 404, 450, 150, 1 \\
		\hline
	\end{tabular}
	\caption{Sizes of neural networks used in the GNN model with the augmented aggregative function first case study.}
	\label{Tab:GNN_arch_1_augmented}
\end{table}

\subsection{Case study two}
\label{sec:case_study_2}

\begin{figure}[H]
    \centering
    \includegraphics[scale=0.7]{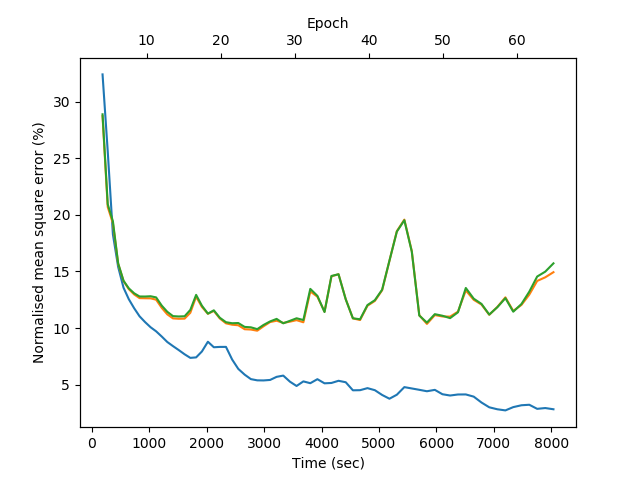}
    \caption{Training history of model using the augmented aggregative function, training (blue), validation (orange) and testing (green) datasets: first case study.}
    \label{fig:augm_aggr_training_history_2}
\end{figure}

Since the use of the augmented aggregative function yielded much better results, in the second problem, it is the only one studied. The training history of this case is shown in Figure \ref{fig:augm_aggr_training_history_2} and the Table \ref{Tab:GNN_arch_2_augmented} contains the architectures of the neural networks used. The validation and testing errors this time were $9.75\%$ and $9.90\%$ respectively. Considering the complexity of the problem it is a satisfactory result. 

\begin{table}[H]
	\centering
	\begin{tabular}{ |p{3cm}||p{3cm}|p{3cm}|p{3cm}|}
		\hline
		 & First CB & Second CB & Third CB\\
		\hline
		Edge NN     & 5, 100, 64 & 234, 100, 64   & 264, 250, 120 \\
		\hline
		Vertices NN & 4, 120, 85  & 298, 200, 100  & 440, 450, 150 \\
		\hline
		Global NN   & 1, 120, 85  & 413, 200, 100  & 640, 450, 150, 1 \\
		\hline
	\end{tabular}
	\caption{Sizes of neural networks used in the GNN model with the augmented aggregative function second case study.}
	\label{Tab:GNN_arch_2_augmented}
\end{table}

In order to further test the algorithm and discover if it has indeed incorporated part of the underlying physics of the problem, another population of trusses was randomly created, this time with number of nodes in the interval $[41, 60]$. Using the model from the second case study to approximate the first natural frequency of the new population, an NMSE of $7.06\%$ was observed. The low error of the later dataset encourages even more the belief that the algorithm indeed encodes the actual physics of the problem.

\section{Conclusions}
\label{sec:conclusions}

In the current paper, a novel way of approximating normal condition characteristics of structures within a population was described. The algorithm is based on the construction of a fibre bundle using a base manifold whose points correspond to structures of a population and as fibres the feature manifolds that characterise each structure. These manifolds are formed by points collected in different potential states in which the structure might be observed. A specific subset of such points, the ones that refer to an undamaged state of the structure are mostly of interest, since they are the normal condition cross section of the bundle.

In order to practically apply the method proposed in \cite{PBSHM6}, a major obstacle was mapping structures into points of a manifold. In the current work, trying to override rather than tackling the specific issue, a machine learning approach to the problem is proposed. The machine learning algorithm used is that of graph neural networks. The specific algorithm allows inference on structures directly, by representing them as graphs. Approaching the problem in such a way, a model was trained using a subset of structures within a population and was validated and tested in two different datasets.

Results reveal that the algorithm is able to approximate efficiently a normal condition characteristic of structures in the population, that of the first natural frequency. Two case studies were presented and in both the algorithm was able to capture the physics of the problem and yield acceptable results. In the second case study, a rather complicated problem, the algorithm was even able to yield good results under the effect of confounding influences (temperature) and even with different types of members, regarding their temperature-stiffness relationship. 

Finally, although GNNs might be considered black-box models, they prove to be located closer to the grey area of the black-white model spectrum. By using inductive biases they are forced to learn the underlying physics of the problem. Inductive biases are introduced through the graphs used as inputs, which define a structure in each data sample rather than just a meaningless one in terms of feature order input vector. In terms of the work presented herein, this is evidenced by the low error the algorithm achieved when its performance was evaluated on a dataset with trusses with more nodes than that on which it was trained on. This last observation could, under some assumptions, be considered as successful extrapolation of the model.

\section{Acknowledgement}
\label{sec:ack}
This project has received funding from the European Unions Horizon 2020 research and innovation programme under the Marie Skodowska-Curie grant agreement No 764547. KW would like to thank the UK Engineering and Physical Sciences Research Council (EPSRC) for an Established Career Fellowship (EP/R003645/1). C. Mylonas and E. Chatzi would further like to gratefully acknowledge the support of the European Research Council via the ERC Starting Grant WINDMIL (ERC-2015-StG \#679843)

\bibliographystyle{unsrt}
\bibliography{imac_21_GT_1}

\end{document}